\documentclass[dvipsnames]{article}
\usepackage{colm2024_conference}

\usepackage{booktabs}
\usepackage{enumitem}
\usepackage{wrapfig}
\usepackage{algorithm}
\usepackage{algpseudocode}
\usepackage{graphicx}
\usepackage{subfigure}
\usepackage[misc]{ifsym}

\usepackage{microtype}
\usepackage{amsmath}
\usepackage{colortbl}
\usepackage[utf8]{inputenc}
\usepackage[T1]{fontenc}
\definecolor{lightgray}{rgb}{0.9,0.9,0.9}
\usepackage{caption}
\usepackage{subcaption}
\usepackage{setspace}
\usepackage{tcolorbox}
\tcbuselibrary{skins}
\usepackage{url}
\usepackage{multirow}
\usepackage{tcolorbox}
\usepackage{tabularx}
\usepackage{blindtext}
\usepackage{pgfplots}
\pgfplotsset{compat=1.18}
\usepackage{tikz}
\usetikzlibrary{er,positioning,bayesnet}
\usepackage{makecell}
\usepackage{tipa}
\usepackage{siunitx}
\usepackage{nicefrac}
\usepackage{listings}
\usepackage{xltabular}
\usepackage{adjustbox}
\usepackage{xurl}
\usepackage{rotating}
\usepackage[normalem]{ulem}

\usepackage{amssymb}
\usepackage{mathtools}
\usepackage{amsthm}
\usepackage{multicol}
\usepackage{inconsolata}
\usepackage[table]{xcolor}
\usepackage{cuted}
\usepackage{capt-of}

\definecolor{myyellow}{rgb}{.99,.94,.82}

\usepackage{booktabs}
\usepackage{multirow}
\usepackage[table]{xcolor}
\usepackage{makecell}

\definecolor{darkgreen}{RGB}{0,140,0}
\definecolor{rowgray}{RGB}{245,245,245}

\definecolor{myboxcolor}{RGB}{245,245,245} 
\definecolor{myframe}{RGB}{0,0,128} 

\newtcolorbox{mybody}{
  colback=myboxcolor,
  colframe=myframe,
  boxrule=1pt, 
  left=1pt,
  right=1pt,
  top=1pt,
  bottom=1pt,
}

\useunder{\uline}{\ul}{}

\usepackage{amsmath,amsfonts,bm}

\def\eqref#1{equation~\ref{#1}}

\def\1{\bm{1}}

\DeclareMathAlphabet{\mathsfit}{\encodingdefault}{\sfdefault}{m}{sl}
\SetMathAlphabet{\mathsfit}{bold}{\encodingdefault}{\sfdefault}{bx}{n}

\newcommand*\justify{%
  \fontdimen2\font=0.4em
  \fontdimen3\font=0.2em
  \fontdimen4\font=0.1em
  \fontdimen7\font=0.1em
  \hyphenchar\font=`\-
}

\renewcommand{\texttt}[1]{%
  \begingroup
  \ttfamily
  \begingroup\lccode`~=`/\lowercase{\endgroup\def~}{/\discretionary{}{}{}}%
  \begingroup\lccode`~=`[\lowercase{\endgroup\def~}{[\discretionary{}{}{}}%
  \begingroup\lccode`~=`.\lowercase{\endgroup\def~}{.\discretionary{}{}{}}%
  \catcode`/=\active\catcode`[=\active\catcode`.=\active
  \justify\scantokens{#1\noexpand}%
  \endgroup
}

\usepackage{makecell}
\usetikzlibrary{tikzmark}
\makeatletter
\newcommand*\myfontsize{%
  \@setfontsize\myfontsize{7}{8}%
}
\makeatother

\definecolor{uclablue}{RGB}{159, 195, 224}

\definecolor{uclagold}{RGB}{255, 240, 180}

\definecolor{aliceblue}{RGB}{255, 238, 241}

\definecolor{cadmiumgreen}{rgb}{0.0, 0.42, 0.24}

\definecolor{myred}{rgb}{0.7, 0.3, 0.0}
\definecolor{myblue}{rgb}{0.2, 0.3, 0.6}
\definecolor{babygreen}{rgb}{0.85, 0.97, 0.85}

\definecolor{purple1}{RGB}{0, 0, 0}
\definecolor{purple2}{RGB}{0, 0, 0}
\definecolor{purple3}{RGB}{0, 0, 0}
\definecolor{purple4}{RGB}{0, 0, 0}

\definecolor{deepblue}{RGB}{48, 58, 82}

\definecolor{deepPurple}{HTML}{000000}

\definecolor{uclablue_old}{rgb}{0, 0, 0}
\hypersetup{
    breaklinks,
    citecolor=citegray,
    colorlinks=true,
    linkcolor=citegray,
    urlcolor=citegray
}

\newcommand{\deltaCell}[1]{%
  \ifdim #1 pt > 0pt
    \textcolor{darkgreen}{#1}%
  \else
    \ifdim #1 pt < 0pt
      \textcolor{red}{#1}%
    \else
      #1%
    \fi
  \fi
}
\newtcolorbox{mybox}[2][]
  {colback = black!5!white, colframe = black!75!black, fonttitle = \bfseries,
    colbacktitle = black!100!black, enhanced, before upper={\fontsize{8}{11}\obeyspaces\obeylines\selectfont}, fontupper=\selectfont,
    attach boxed title to top left={yshift=-2.2mm,xshift=4mm},
    title=#2,#1}

\title{BabyVision: Visual Reasoning Beyond Language}

\author{%
\small{Liang Chen$^{1,6}\thanks{Equal Core Contributors. Correspondence: Liang Chen <liangchen@unipat.ai>, Kuan Li <kuanli@unipat.ai>}$, Weichu Xie$^{1,6*}$, Yiyan Liang$^{1,6*}$, Hongfeng He$^{1*}$, Hans Zhao$^{1*}$, Zhibo Yang$^{3}$, Zhiqi Huang$^{4}$, Haoning Wu$^{4}$, Haoyu Lu$^{4}$, Y. charles$^{4}$, Yiping Bao$^{4}$, Yuantao Fan$^{5}$, Guopeng Li$^{5}$, Haiyang Shen$^{1,6}$, Xuanzhong Chen$^{1,7}$, Wendong Xu$^{1}$, Shuzheng Si$^{7}$, Zefan Cai$^{8}$, Wenhao Chai$^{9}$, Ziqi Huang$^{10}$, Fangfu Liu$^{7}$, Tianyu Liu$^{6}$, Baobao Chang$^{6}$, Ming Wu$^{11}$, Xiaobo Hu$^{2}$, Kaiyuan Chen$^{2}$, Yixin Ren$^{2}$, Yang Liu$^{2}$, Yuan Gong$^{2}$, Kuan Li$^{1}$}
  \\[0.8em]
  {\fontsize{10pt}{11pt}\selectfont          
  $^1$UniPat AI $^2$xbench $^3$Alibaba Group $^4$MoonShot AI $^5$StepFun $^6$Peking University $^7$Tsinghua University $^8$University of Wisconsin--Madison $^9$Princeton University $^{10}$Nanyang Technological University $^{11}$0G Labs}\\%
}

\begin{document}

 \maketitle


\newcommand{\datasetname}{\textsc{BabyVision}\xspace}
\newcommand{\datasetnamegen}{\textsc{BabyVision-Gen}\xspace}

\begin{abstract}

While humans develop core visual skills long before acquiring language, contemporary Multimodal LLMs (MLLMs) still rely heavily on linguistic priors to compensate for their fragile visual understanding. We uncovered a crucial fact: state-of-the-art MLLMs consistently fail on basic visual tasks that humans, even 3-year-olds, can solve effortlessly. To systematically investigate this gap, we introduce \datasetname, a benchmark designed to assess core visual abilities independent of linguistic knowledge for MLLMs.
\datasetname spans a wide range of tasks, with 388 items divided into 22 subclasses across four key categories. Empirical results and human evaluation reveal that leading MLLMs perform significantly below human baselines. Gemini3-Pro-Preview scores 49.7, lagging behind 6-year-old humans and falling well behind the average adult score of 94.1. These results show despite excelling in knowledge-heavy evaluations, current MLLMs still lack fundamental visual primitives. Progress in \datasetname represents a step toward human-level visual perception and reasoning capabilities. We also explore solving visual reasoning with generation models by proposing \datasetnamegen and automatic evaluation toolkit. Our code and benchmark data are released at \href{https://github.com/UniPat-AI/BabyVision}{\github~UniPat-AI/BabyVision} for reproduction.
\end{abstract}


\begin{figure}[h]
    \centering
    \includegraphics[width=0.8\linewidth]{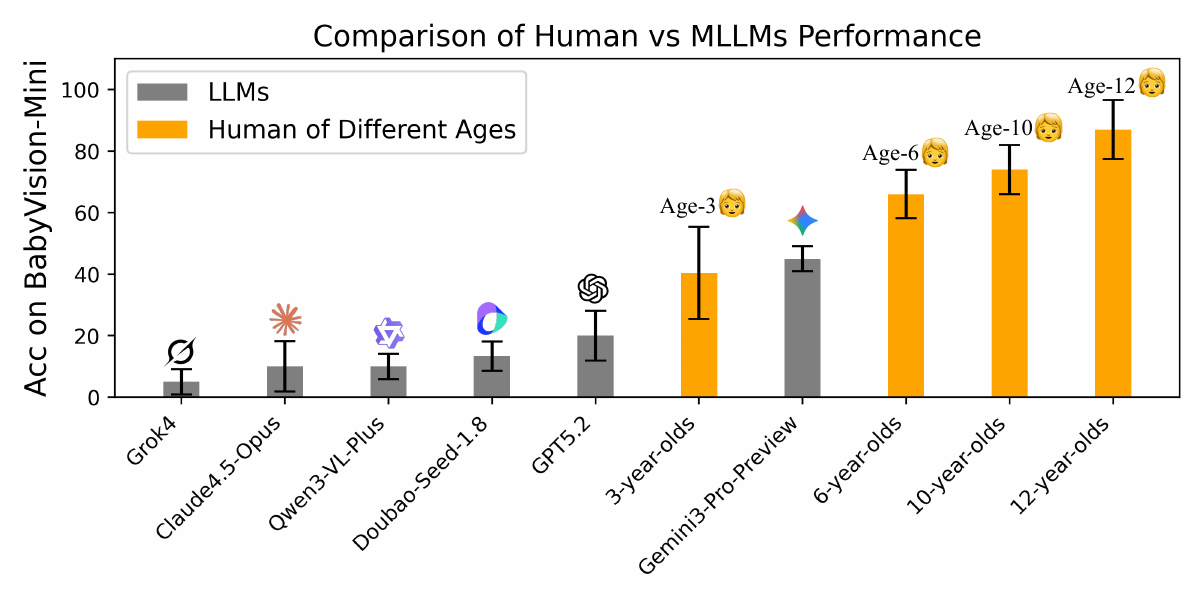}
    \caption{Performance on \datasetname among MLLMs and human of different ages.}
    \label{fig:abs_fig}
\end{figure}

\newpage

\section{Introduction}

\begin{figure}[h]
    \centering
    \includegraphics[width=0.8\linewidth]{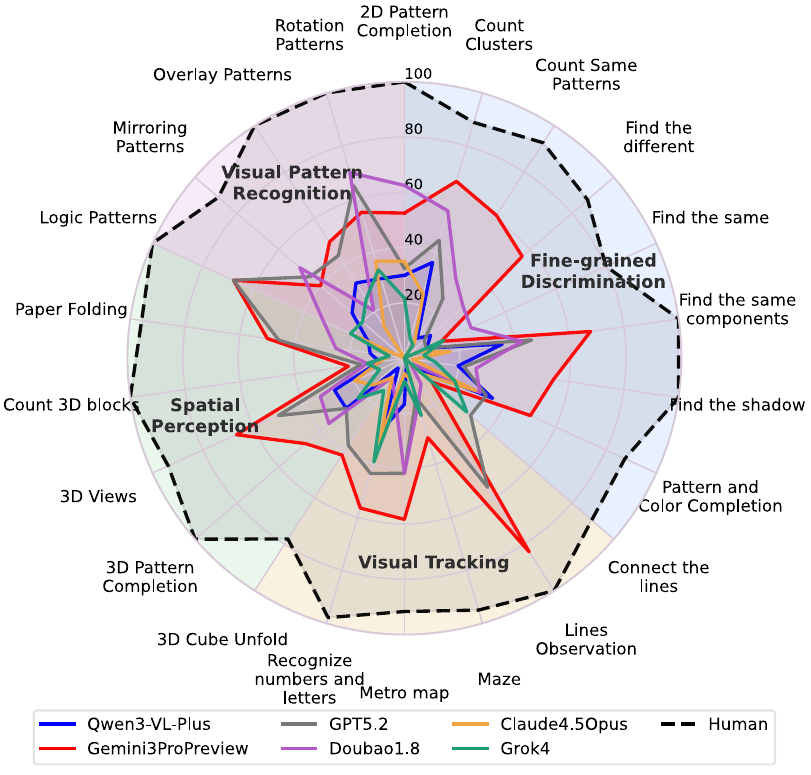}
    \caption{Fine-grained performance analysis on the full \datasetname benchmark.}
    \label{fig:abs_full_benchmark}
\end{figure}

Human visual understanding emerges remarkably early in life—well before children acquire language or formal symbolic reasoning. Within the first months, infants can already discriminate shapes and textures, track moving objects, infer occlusion and depth, and form expectations about simple physical events~\citep{johnson2010development,braddick2011development,kellman2006infant}.

These \emph{early-vision abilities} form a foundation for later cognitive functions—including language, abstract reasoning, and motor planning—by supporting structured representations of the physical world.

In contrast, contemporary Multimodal Large Language Models (MLLMs)~\citep{team2023gemini,GPT-5.2,bai2025qwen3vltechnicalreport} appear to exhibit an inverted profile of competence. They achieve strong performance on many high-level, knowledge-intensive benchmarks that typically demand substantial education or domain expertise (e.g., HLE~\citep{phan2025humanitysexam}, MMMU~\citep{yue2024mmmu}), including multi-step geometric and mathematical reasoning (e.g., MathVista~\citep{lu2024mathvista}, MathVision~\citep{wang2024mathvision}), large-scale recognition of people and places (e.g., MME~\citep{fu2024mme}), and even medically styled question answering (e.g., DrVD-Bench~\citep{zhou2025drvd}). Yet, despite these successes, our studies reveal a consistent weakness in \emph{basic} visual tasks that children (ages 3–12) can solve with little or no language mediation—such as visual discrimination, visual tracking, spatial understanding, and simple visual pattern recognition. Even the strongest model we tested, Gemini3-Pro-Preview, remains approximately 20\% behind 6-year-old children in our pilot study, as shown in Figure~\ref{fig:abs_fig}. This discrepancy suggests that current MLLMs still lack the atomic visual competencies that underlie human vision from its earliest stages. However, existing evaluations rarely target these \textbf{beyond-language visual understanding} abilities in a systematic way.

To fill this gap, we introduce \datasetname, a benchmark designed with a scientific and rigorous data curation pipeline to probe the atomic visual skills humans develop early in life. \datasetname aims to minimize reliance on linguistic knowledge by emphasizing tasks driven by perception and pattern regularities rather than textual reasoning or semantic priors. The benchmark contains 388 unique questions spanning 22 subclasses across four domains: \emph{Fine-grained Discrimination}, \emph{Visual Tracking}, \emph{Spatial Perception}, and \emph{Visual Pattern Recognition}. Together, these domains cover a broad spectrum of early-vision competencies with substantial diversity in visual conditions and task structure.

\begin{figure}[t] \centering \includegraphics[width=1\textwidth]{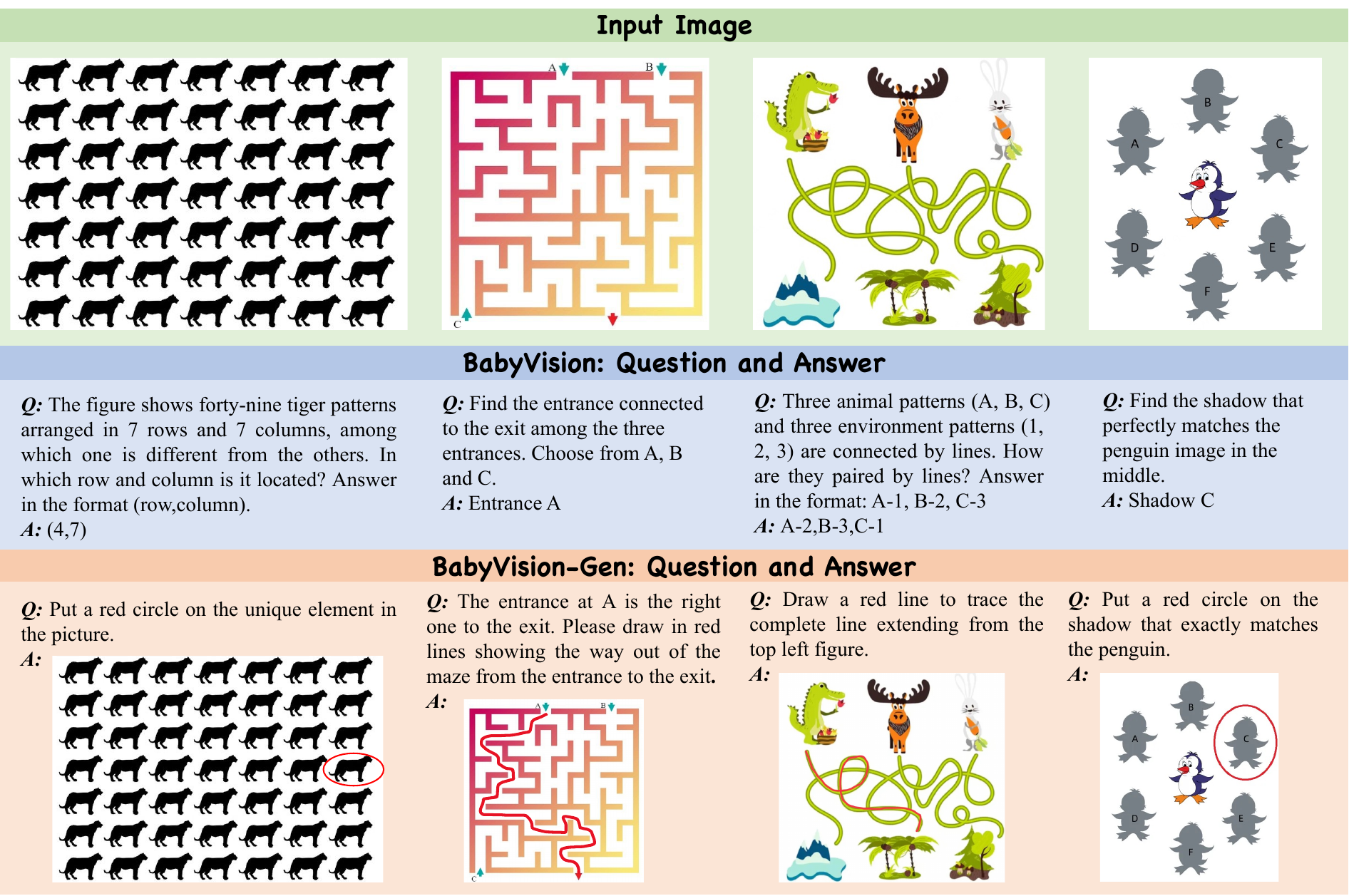} \caption{Examples of \datasetname and \datasetnamegen. While \datasetname evaluates visual understanding through language output, \datasetnamegen evaluates visual reasoning through image generation.} \label{fig:babyvision-gen-comparison} \end{figure}

As shown in Figure~\ref{fig:abs_full_benchmark}, we evaluate state-of-the-art MLLMs on \datasetname, covering both proprietary and open models (e.g., Gemini3-Pro-Preview, GPT-5.2, Qwen3VL-235B-Thinking). In general, we observe a substantial gap between current MLLMs and human performance on early-vision competencies. Specifically, the best-performing model achieves an overall accuracy of \textbf{49.7}\%, whereas adult human testers reach \textbf{94.1}\% (a \textbf{44.4}\% absolute gap), with consistent deficits across all four domains. The largest failures appear in \emph{Visual Tracking} and \emph{Spatial Perception}, where models frequently exhibit errors such as a \emph{loss of manifold identity} (e.g., losing track of curves through intersections) and a \emph{failure of spatial imagination} (e.g., the inability to mentally transform 3D structures). These results indicate that strong high-level multimodal reasoning does not imply robust foundational visual competence, and that \datasetname exposes a distinct failure mode not captured by existing benchmarks.

Beyond textual evaluation, we observe that many questions from \datasetname are most naturally solved visually: humans typically draw on the image—by tracing paths, completing patterns, or marking spatial relations—rather than verbalizing their reasoning. Motivated by this, we introduce \datasetnamegen, a generative counterpart to \datasetname that evaluates visual reasoning through visual generation \textbf{beyond language output}. We also develop an automatic evaluation tool achieving a 96\% agreement rate with human judgments. Experiments with frontier image and video generators (e.g., Nano-Banana-Pro, Sora-2) show promising gains on tasks that are challenging for MLLMs, especially visual tracking and fine-grained discrimination, though overall reliability and solution consistency remain limited.

In summary, this paper makes three main contributions: (1) We introduce \datasetname, a refined benchmark for evaluating \textbf{beyond-language} visual reasoning abilities, containing \textbf{388} questions across \textbf{22} subclasses in four domains. (2) Motivated by the observation that many tasks are most naturally solved via visual generation, we further propose \datasetnamegen, an extension of \datasetname that evaluates visual reasoning ability for generation models \textbf{beyond language output}, along with a reliable automatic evaluation toolkit. (3) We conduct a comprehensive evaluation of state-of-the-art MLLMs and generative models, revealing a new performance landscape marked by a substantial human–model gap (44.4\%) and pronounced inter-model disparities (e.g., Gemini3-Pro-Preview outperforming the runner-up by approximately 15 points). We further provide fine-grained analyses of failure patterns at both the domain and subclass levels, and investigate how RLVR training and generative modeling contribute to improved visual reasoning performance.
\section{Related Work}
\subsection{Benchmarks for Multimodal Large Language Models}
The rapid progress of Multimodal Large Language Models (MLLMs) has driven the development of diverse evaluation benchmarks. A prominent line of work targets expert-level, knowledge-intensive multimodal reasoning. MMMU~\citep{yue2024mmmu} assembles 11.5K college-level questions spanning 30 disciplines, while HLE~\citep{phan2025humanitysexam} pushes further toward ``humanity's last exam'' with expert-level problems. Mathematical reasoning benchmarks such as MathVista~\citep{lu2024mathvista}, MathVision~\citep{wang2024mathvision}, and MathVerse~\citep{zhang2024mathverse} focus on geometric and quantitative reasoning grounded in images, while MME~\citep{fu2024mme} and MMEvalPro~\citep{MMEvalPro} provide comprehensive evaluations across perception and cognition dimensions. On these benchmarks, leading models achieve impressive results---Gemini 3 Pro attains 92.8\% on MMMU---demonstrating the increasingly effective integration of visual inputs with domain knowledge.

However, strong results on semantic and knowledge-driven benchmarks do not necessarily translate into robust visual perception. This inverted competence profile---where models excel at expert-level tasks but struggle with basic perception---has been documented across several studies. BLINK~\citep{fu2024blink} reveals a substantial gap between humans (95.7\%) and leading MLLMs on classical vision problems that require no specialized knowledge. MMStar~\citep{chen2024mmstar} demonstrates that models can achieve 42.9\% on MMMU \emph{without} visual input, exposing severe language leakage and over-reliance on textual shortcuts. MMVP~\citep{mmvp} further probes visual limitations and finds that models fail to distinguish images with clear perceptual differences. These findings reveal a recurring blind spot: existing benchmarks predominantly draw from tasks designed for adult experts and rely on semantic recognition rather than perceptual primitives. \datasetname addresses this gap by targeting pre-linguistic visual abilities---the foundational competencies humans develop before language acquisition---and comparing model performance directly to children aged 3--12.

\subsection{Early Vision and Developmental Foundations}
The design of \datasetname draws on developmental psychology research establishing that humans acquire core visual competencies well before language. Infants demonstrate object permanence by 3--4 months~\citep{baillargeon1985object}, track objects through occlusion~\citep{johnson2010development}, and discriminate depth and shape before acquiring language~\citep{kellman2006infant,braddick2011development}. The core knowledge hypothesis~\citep{spelke2000core} posits innate systems for representing objects, space, numbers, and agents---capacities that are independent of linguistic mediation. By ages 3--6, children perform complex visual discrimination, understand spatial transformations, and recognize patterns, often demonstrating competence through non-verbal responses~\citep{karmiloff1994beyond,golomb2004child}.

This developmental trajectory motivates key design choices in \datasetname. We select tasks solvable by young children, ensuring they test foundational rather than expert-level abilities. We minimize verbal reasoning demands by focusing on perceptual regularities rather than semantic knowledge. We compare model performance directly to children at different developmental stages, revealing not just whether models fail, but \emph{how far} they fall below human visual development. Developmental research also shows that children externalize visual reasoning through drawing before verbalizing solutions~\citep{kellogg1969analyzing}---they trace paths, complete patterns, and mark spatial relations rather than describing them. This observation motivates \datasetnamegen: evaluating models solely on verbal answers may underestimate visual competence that could be demonstrated through visual output, suggesting generation-based evaluation as a complementary paradigm.
\section{\datasetname}

\label{sec:BabyVision}

In this section, we present \datasetname, a benchmark designed to evaluate the early-vision capabilities of Multimodal Large Language Models (MLLMs). Unlike previous benchmarks that focus on complex semantic reasoning or domain knowledge, \datasetname targets foundational visual skills that humans acquire during early development—abilities that emerge before or alongside language acquisition. We split vision-centric reasoning into four core categories: \emph{Fine-grained Discrimination} (detecting subtle visual differences; 8 subtypes), \emph{Visual Tracking} (following paths, lines, and trajectories; 5 subtypes), \emph{Spatial Perception} (understanding 3D structures and relationships; 5 subtypes), and \emph{Visual Pattern Recognition} (identifying logical and geometric patterns; 4 subtypes). Together, these comprise 22 basic subtypes, each targeting a specific fundamental visual capability. Through a rigorous data curation pipeline, we construct 388 questions spanning a wide diversity of visual reasoning tasks.

\subsection{Data Curation Pipeline}
\label{subsec:annotation_pipeline}

\begin{figure}[h]
\centering
\includegraphics[width=\linewidth]{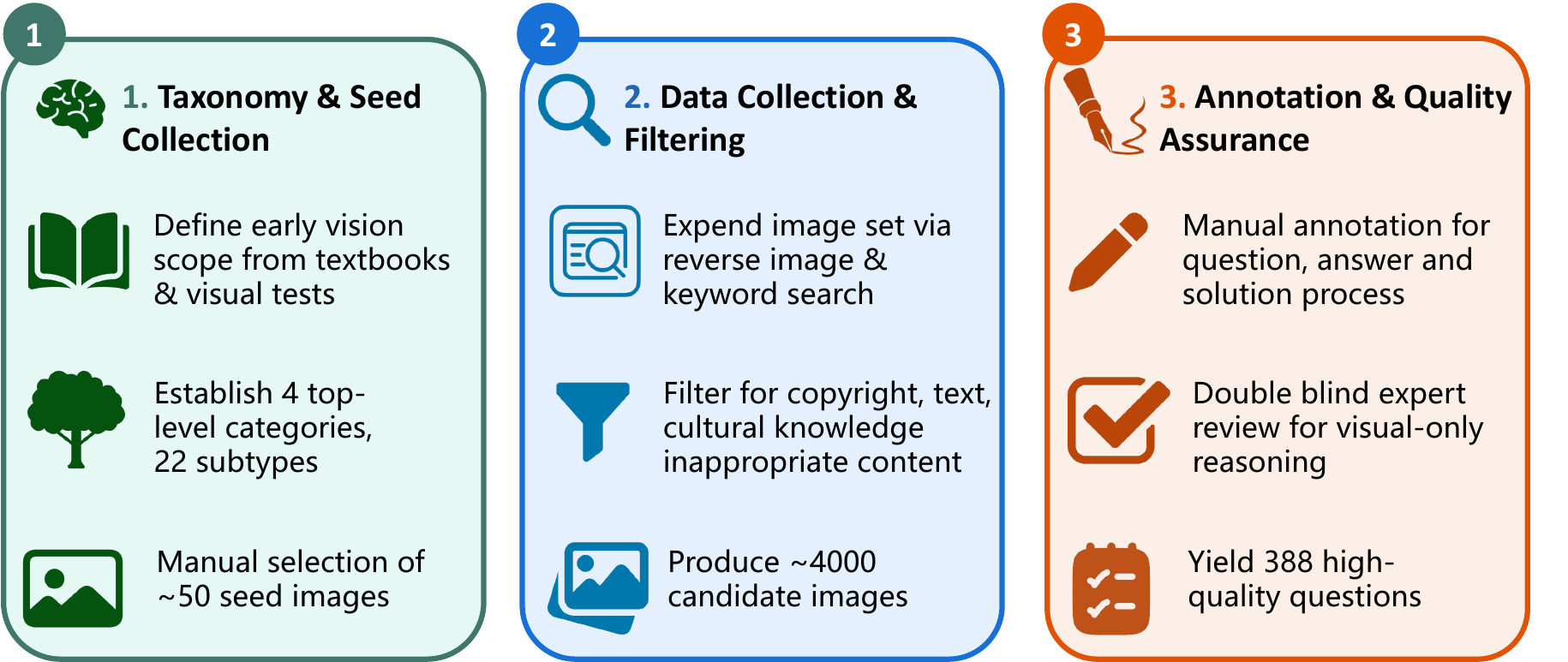}
\caption{Overview of the multi-stage data collection and curation pipeline for \datasetname: taxonomy design \& seed selection, data augmentation \& filtering, and annotation \& quality assurance.}
\label{fig:data_collection}
\end{figure}

To ensure high quality and validity, we employ a rigorous multi-stage data collection and curation process, as shown in Figure~\ref{fig:data_collection}. The whole process includes the following three steps:

\paragraph{Taxonomy Definition and Seed Collection.}
We first define the scope of ``early vision'' grounded in developmental psychology. Consulting textbooks designed for children under 12 and standardized visual development tests, we identify core visual competencies and establish four top-level categories with 22 specific subtypes. For each subtype, we manually select 4--5 ``seed images'' that exemplify the visual task, yielding approximately 100 high-quality seed examples that serve as prototypes for data expansion.

\begin{figure}[!t]
\centering
\includegraphics[width=\textwidth]{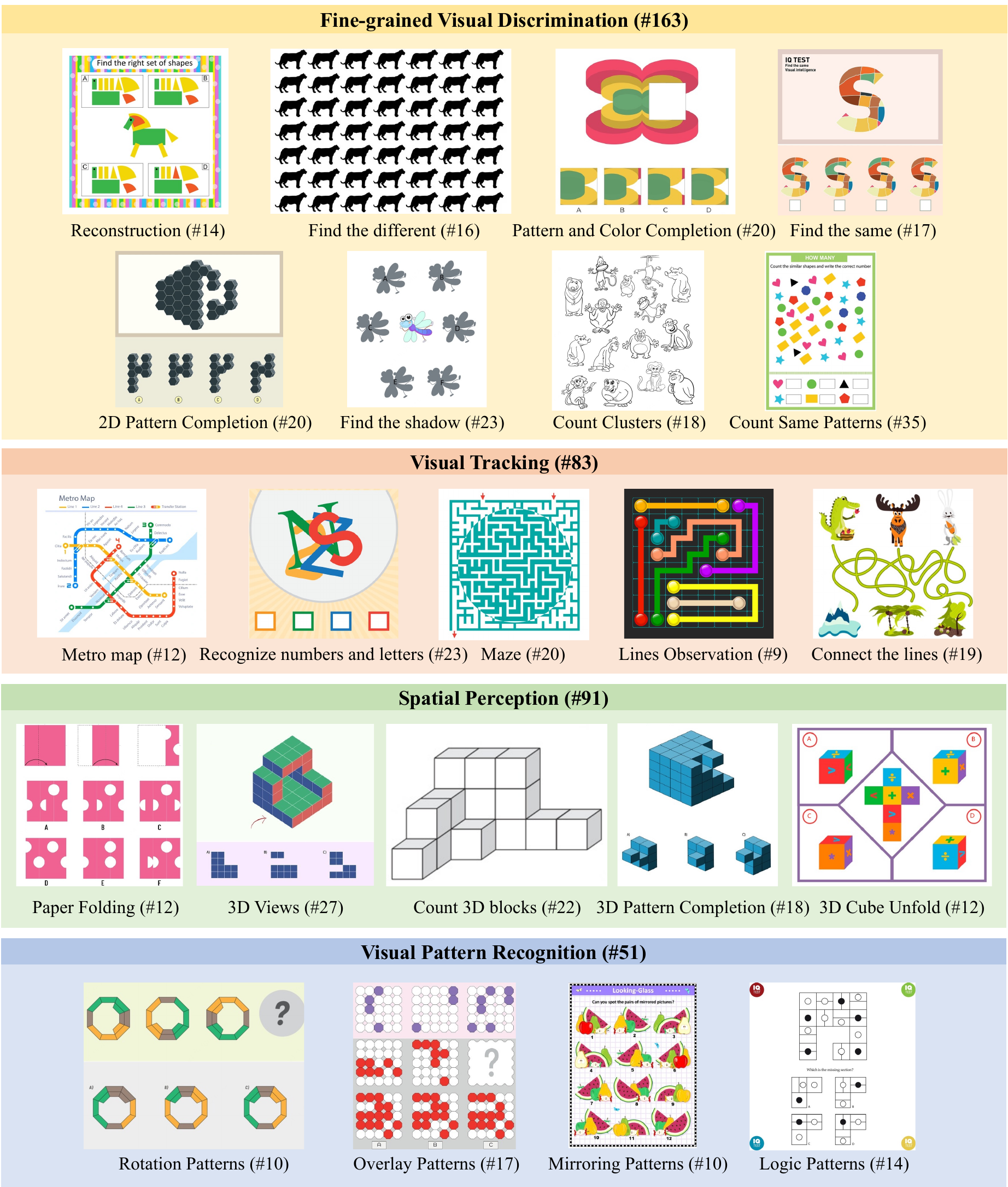}
\caption{Example questions and the number of examples (\#) from \datasetname spanning the four core categories and 22 types.}
\label{fig:display_example}
\end{figure}

\paragraph{Data Collection and Filtering.}
Using the seed images, we expand the dataset through reverse image search and keyword-based retrieval, crawling similar images from the internet. We strictly adhere to copyright regulations, retaining only images permitted for academic use. Images containing substantial text, requiring cultural knowledge, or depicting inappropriate content are filtered out. This stage produces approximately 4,000 candidate images.

\paragraph{Annotation and Quality Assurance.}
Each candidate image undergoes manual annotation by trained annotators. Annotators first determine whether the image can support a meaningful visual question aligned with our taxonomy—images that rely heavily on text reading or require cultural background knowledge are discarded. For valid images, annotators write specific questions and answers and, crucially, provide a detailed ``solution process'' proving the answer's correctness.

All annotations then pass through a double-blind review: two independent experts verify that the answer is unambiguous and derivable largely from visual analysis rather than language. A question is included only if both experts agree on the answer and reasoning logic. Disagreements are returned to the original annotator for revision; questions that remain contentious after revision are permanently discarded.

\subsection{Dataset Statistics}
\label{subsec:dataset_statistics}

The final \datasetname benchmark comprises 388 carefully curated questions, each paired with a unique image. The dataset includes 135 multiple-choice questions (34.8\%) and 253 fill-in-the-blank questions (65.2\%), with balanced answer distributions to mitigate position bias. The questions are concise (averaging 25.9 words) and tightly grounded in visual content, minimizing opportunities for language-based shortcuts. The distribution of task categories is shown in Figure~\ref{fig:display_example}. In addition, we collect 1,400 training examples following the same construction pipeline but sourced more broadly, enabling an investigation of how training influences model performance on \datasetname, which is discussed in Section~\ref{subsec:training}.

\subsection{Evaluation}
\label{subsec:evaluation_vlm}

For inference, we use a unified prompt template across all models: \texttt{{question} Think about the question and give your final answer in  format.} This encourages models to reason through the problem before answering and facilitates automated answer extraction. By default, we select the highest reasoning effort of the tested models if adjustable.

We evaluate model responses using an LLM-as-judge approach. Given a model's output answer and the ground-truth answer, we query Qwen3-Max to determine semantic equivalence, which shows 100\% consistency with human evaluators. The full judge prompt is provided in Appendix~\ref{appx:eval}.

\section{\datasetnamegen}
\label{sec:BabyVision-Gen}

A key observation from \datasetname is that many tasks admit solutions that are most naturally expressed \emph{visually}. When humans solve maze navigation or pattern completion problems, they often draw trajectories or complete shapes rather than verbalizing step-by-step reasoning~\citep{kellogg1969analyzing, golomb2004child}. Similarly, in find-the-difference tasks, solutions are typically conveyed by directly circling the differing regions in the image. This observation motivates \datasetnamegen, a generative extension that evaluates whether models can perform visual reasoning through image (or video) generation, providing an alternative pathway that bypasses the verbalization bottleneck.

\subsection{Test Reasoning in Generation Models}

\datasetnamegen adapts a subset of \datasetname questions for visual output and alters the prompt to support generation. For each question, we provide a generation prompt instructing the model to produce an image that demonstrates the solution—tracing a path, completing a pattern, or marking spatial relationships. This formulation tests whether visual generation models can leverage visual reasoning capabilities that may exceed their ability to verbalize answers. We show some examples of \datasetnamegen in Figure~\ref{fig:babyvision-gen-comparison}.

\paragraph{Statistics.} \datasetnamegen comprises 280 questions across 21 subtypes organized into the same four categories as \datasetname: Fine-grained Discrimination (128 questions, 8 subtypes), Visual Tracking (55 questions, 4 subtypes), Spatial Perception (59 questions, 5 subtypes), and Visual Pattern Recognition (38 questions, 4 subtypes). Note that one subtype from \datasetname is excluded as it does not naturally admit a generative solution format. Each instance includes the original image, a generation prompt (averaging 22.9 words), and a reference solution image for evaluation, reflecting the consensus of three human annotators. The detailed distribution across categories and subtypes is provided in Table~\ref{tab:gen_statistics} in Appendix~\ref{appx:gen-stats}.

\subsection{Evaluation}
\label{subsec:evaluation_gen}

Evaluating \datasetnamegen requires determining whether a model’s generated image expresses the correct visual solution. We standardize inference by instructing models to annotate the original input image with minimal overlays (e.g., circles, lines, arrows, or text), preserving the underlying visual context while making the predicted answer explicit.

For automatic evaluation, we employ Gemini-3-Flash as a judge. The judge is given three images: (i) the input image, (ii) a human-annotated ground-truth solution, and (iii) the model-generated output. It returns a binary decision indicating whether the generated annotation matches the ground truth under subtype-specific criteria (e.g., identical selected option; same traced route for mazes). Full prompts and criteria are provided in Appendix~\ref{appx:eval-gen}.

To further assess the reliability of automatic evaluation, we perform human evaluation on all Nano-Banana-Pro outputs using PhD-level annotators. Automatic and human judgments agree on \textbf{96.1\%} of instances (269/280; F1=0.924), supporting the use of the automatic judge for scalable evaluation. Subtype-wise agreement rates and the confusion matrix are reported in Appendix~\ref{appx:eval-gen}.

\section{Main Experiments}
In this section, we present a comprehensive evaluation on the BabyVision benchmarks. We first describe the experimental setup in Section~\ref{sec: Experiment Setup}, followed by the results on \datasetname in Section~\ref{sec:overall_results} and on \datasetnamegen in Section~\ref{sec:gen_results}.

\subsection{Evaluation Setup}
\label{sec: Experiment Setup}

\paragraph{Models.} Our evaluation encompasses 11 frontier models, comprising 5 proprietary and 6 open-source systems. The proprietary models include Gemini3-Pro-Preview~\citep{team2023gemini}, GPT-5.2~\citep{GPT-5.2}, Qwen3-VL-Plus~\citep{bai2025qwen3vltechnicalreport}, Doubao-1.8~\citep{Seed1.6}, and Claude-4.5-Opus~\citep{Claude-Opus-4.5}. The open-source selection consists of Qwen3VL-235B-Thinking~\citep{bai2025qwen3vltechnicalreport}, InternVL3.5-241B~\citep{wang2025internvl35advancingopensourcemultimodal}, Step3~\citep{stepfun2025step3largeaffordablemodelsystem}, KimiVL-A3B~\citep{kimiteam2025kimivltechnicalreport}, MimoVL-7B-RL~\citep{coreteam2025mimovltechnicalreport}, and GLM4.6V~\citep{GLM-4.6V}.

Furthermore, to examine the influence of model scaling and reasoning modes (thinking vs. non-thinking), we evaluate additional variants of the Qwen3VL family, including Qwen3VL-235B (Thinking/Instruct) and the smaller Qwen3VL-32B/8B/4B-Thinking models.

\paragraph{Evaluation Settings.}

We report the Avg@3 results for all evaluated models. For proprietary models, we utilize the official APIs without modifying default parameters, with the exception of reasoning effort, which is set to the maximum level where adjustable. For open-source models, we adhere to the inference temperatures recommended in their official papers; unless otherwise specified, we employ a default sampling temperature of 1.0. The inference and evaluation prompts for \datasetname and \datasetnamegen are detailed in Section~\ref{subsec:evaluation_vlm} and Section~\ref{subsec:evaluation_gen}, respectively.

\paragraph{Human Testers.}

In our primary experiment, we conducted a comparative study using 20 representative samples (BabyVision-Mini) spanning diverse categories. We recruited participants across four age groups—3, 6, 10, and 12 years old—with 20 individuals per group\footnote{Testers aged 3-12 are from a single school population. Results may vary across different populations, educational backgrounds, or cultural contexts.}. The study was conducted with full permission from school authorities and parents, as well as the voluntary assent of the participants. Each group was allotted one class period (45 minutes) to complete the test. In our comprehensive evaluation of \datasetname, 16 adult participants, each holding at least a bachelor's degree, completed the entire benchmark (388 questions). Human testers are not included in the evaluation of \datasetnamegen.

\subsection{\datasetname Results}
\label{sec:overall_results}

\paragraph{Proprietary Model Performance.}

The strongest proprietary model performance remains substantially below the human baseline (94.1\%). Among proprietary systems, Gemini3-Pro-Preview achieves the highest overall score (49.7\%), followed by GPT-5.2 (34.4\%) and Doubao-1.8 (30.2\%), while the remaining models trail by a wide margin (e.g., Qwen3-VL-Plus at 19.2\%, Grok-4 at 16.2\%, Claude-4.5-Opus at 14.2\%). The large gap to human performance is consistent across all four task families, with no single category dominating the error profile. This pattern suggests that current systems still lack foundational visual competencies, pointing to a systemic limitation rather than an isolated weakness.

At the subtype level, several tasks prove challenging for nearly all systems. In particular, \emph{Count 3D Blocks} yields uniformly low accuracy (best: 20.5\%), indicating that compositional 3D counting under occlusion and viewpoint variation remains difficult. Similarly, \emph{Find the Same} reaches only 26.5\% at best (Doubao-1.8), despite a human baseline of 80\%. Such failures are informative because they reflect deficiencies in structured scene representations (e.g., object permanence and depth-aware composition), rather than shortcomings in recognition or superficial pattern matching.

\begin{table*}[!t]
\renewcommand\arraystretch{1}
\setlength{\aboverulesep}{1.3pt}
\setlength{\belowrulesep}{1.3pt}
\centering
\resizebox{\linewidth}{!}{
\begin{tabular}{c | l |c| c c | c c | c c | c c | c c| c c}
\toprule
{\multirow{2}{*}{\textbf{Type}}} &
{\multirow{2}{*}{\textbf{Sub-Type}}} &
{\multirow{2}{*}{\textbf{Human}}} &
\multicolumn{2}{c|}{\textbf{Gemini3-Pro-Preview}} &
\multicolumn{2}{c|}{\textbf{GPT-5.2}} &
\multicolumn{2}{c|}{\textbf{Doubao-1.8}} &
\multicolumn{2}{c|}{\textbf{Qwen3-VL-Plus}} &
\multicolumn{2}{c|}{\textbf{Claude-4.5-Opus}}&
\multicolumn{2}{c}{\textbf{Grok-4}} \\
\cmidrule(lr){4-5} \cmidrule(lr){6-7} \cmidrule(lr){8-9} \cmidrule(lr){10-11} \cmidrule(lr){12-13} \cmidrule(lr){14-15}
& & &
\textbf{Avg ($\mu$) $\uparrow$} & \textbf{Std ($\sigma$) $\downarrow$} &
\textbf{Avg ($\mu$) $\uparrow$} & \textbf{Std ($\sigma$) $\downarrow$} &
\textbf{Avg ($\mu$) $\uparrow$} & \textbf{Std ($\sigma$) $\downarrow$} &
\textbf{Avg ($\mu$) $\uparrow$} & \textbf{Std ($\sigma$) $\downarrow$} &
\textbf{Avg ($\mu$) $\uparrow$} & \textbf{Std ($\sigma$) $\downarrow$} &
\textbf{Avg ($\mu$) $\uparrow$} & \textbf{Std ($\sigma$) $\downarrow$} \\
\midrule
\multirow{10}{*}{\makecell{Fine-grained \\ Discrimination}}
& 2D Pattern Completion & 100 & 52.5 & 2.5 & 32.5 & 7.5 & \textbf{62.5} & 2.5 & 30.0 & 5.0 & 35.0 & 4.1 & 21.7 & 4.7 \\
& Count Clusters & 88.9 & \textbf{66.7} & 0.0 & 44.4 & 5.6 & 55.6 & 5.6 & 36.1 & 2.8 & 24.1 & 2.6 & 7.4 & 2.6 \\
& Count Same Patterns & 92.7 & \textbf{61.4} & 4.3 & 25.7 & 2.9 & 34.3 & 5.7 & 8.6 & 2.9 & 6.7 & 1.4 & 5.7 & 0.0 \\
& Find the Different & 87.5 & \textbf{56.3} & 6.3 & 9.4 & 3.1 & 28.1 & 3.1 & 12.5 & 0.0 & 0.0 & 0.0 & 0.0 & 0.0 \\
& Find the Same & 80 & 14.7 & 2.9 & 8.8 & 2.9 & \textbf{26.5} & 2.9 & 8.8 & 8.8 & 5.9 & 0.0 & 15.7 & 2.8 \\
& Reconstruction & 100 & \textbf{67.9} & 3.6 & 46.4 & 3.6 & 42.9 & 0.0 & 35.7 & 0.0 & 16.7 & 3.4 & 7.1 & 5.8 \\
& Find the Shadow & 100 & \textbf{54.4} & 6.5 & 21.7 & 0.0 & 26.1 & 4.4 & 19.6 & 10.9 & 2.9 & 2.1 & 11.6 & 7.4 \\
& Pattern and Color Completion & 87.5 & \textbf{50.0} & 10.0 & 32.5 & 7.5 & 30.0 & 5.0 & 35.0 & 0.0 & 28.3 & 2.4 & 20.0 & 4.1 \\
\cmidrule(lr){2-15}
& \cellcolor{blue!5}\textbf{Sub-Type Overall Results} & \cellcolor{blue!5}92.3
& \cellcolor{blue!5}\textbf{46.2} &  \cellcolor{blue!5}5.5
& \cellcolor{blue!5}27.3 & \cellcolor{blue!5}0.9
& \cellcolor{blue!5}39.2 & \cellcolor{blue!5}0.0
& \cellcolor{blue!5}21.8 & \cellcolor{blue!5}2.2
& \cellcolor{blue!5}14.3 & \cellcolor{blue!5}0.3
& \cellcolor{blue!5}11.0 & \cellcolor{blue!5}2.7 \\
\midrule
\multirow{6}{*}{\makecell{Visual Tracking}}
& Connect the Lines & 89.5 & 13.2 & 2.6 & \textbf{31.6} & 0.0 & 2.6 & 2.6 & 5.3 & 5.3 & 8.8 & 6.6 & 29.8 & 2.5 \\
& Lines Observation & 100 & \textbf{83.3} & 5.6 & 55.6 & 0.0 & 11.1 & 11.1 & 0.0 & 0.0 & 0.0 & 0.0 & 0.0 & 0.0 \\
& Maze & 95.0 & \textbf{30.0} & 5.0 & 15.0 & 0.0 & 17.5 & 2.5 & 2.5 & 2.5 & 8.3 & 2.4 & 21.7 & 4.7 \\
& Metro Map & 91.7 & \textbf{58.3} & 8.3 & 41.7 & 0.0 & 41.7 & 8.3 & 16.7 & 0.0 & 5.6 & 3.9 & 8.3 & 6.8 \\
& Recognize Numbers and Letters & 97.8 & \textbf{56.5} & 4.4 & 43.5 & 0.0 & 13.0 & 0.0 & 26.1 & 4.4 & 31.9 & 2.1 & 39.1 & 0.0 \\
\cmidrule(lr){2-15}
& \cellcolor{blue!5}\textbf{Sub-Type Overall Results} &\cellcolor{blue!5}94.6
& \cellcolor{blue!5}\textbf{43.4} & \cellcolor{blue!5}2.4
& \cellcolor{blue!5}34.9 & \cellcolor{blue!5}0.0
& \cellcolor{blue!5}15.7 & \cellcolor{blue!5}1.2
& \cellcolor{blue!5}11.5 & \cellcolor{blue!5}0.6
& \cellcolor{blue!5}13.7 & \cellcolor{blue!5}2.8
& \cellcolor{blue!5}24.1 & \cellcolor{blue!5}1.0 \\
\midrule
\multirow{6}{*}{\makecell{Spatial Perception}}
& 3D Cube Unfold & 77.8 & \textbf{41.7} & 8.3 & 37.5 & 4.2 & 8.3 & 0.0 & 4.2 & 4.2 & 8.3 & 6.8 & 13.9 & 10.4 \\
& 3D Pattern Completion & 100.0 & \textbf{47.2} & 13.9 & 27.8 & 0.0 & 36.1 & 8.3 & 27.8 & 5.6 & 14.8 & 2.6 & 22.2 & 7.9 \\
& 3D Views & 93.8 & \textbf{66.7} & 7.4 & 50.0 & 1.9 & 33.3 & 3.7 & 27.8 & 5.6 & 19.8 & 6.3 & 9.9 & 4.6 \\
& Count 3D Blocks &100 & \textbf{20.5} & 2.3 & 15.9 & 6.8 & 13.6 & 4.6 & 9.1 & 4.6 & 9.1 & 3.7 & 13.6 & 3.7 \\
& Paper Folding &95.8 & \textbf{50.0} & 0.0 & 45.8 & 12.5 & 25.0 & 0.0 & 12.5 & 4.2 & 5.6 & 3.9 & 5.6 & 3.9 \\
\cmidrule(lr){2-15}
& \cellcolor{blue!5}\textbf{Sub-Type Overall Results} &\cellcolor{blue!5}94.7
& \cellcolor{blue!5}\textbf{53.7} & \cellcolor{blue!5}1.5
& \cellcolor{blue!5}35.2 & \cellcolor{blue!5}2.2
& \cellcolor{blue!5}24.7 & \cellcolor{blue!5}0.6
& \cellcolor{blue!5}18.1 & \cellcolor{blue!5}1.7
& \cellcolor{blue!5}12.8 & \cellcolor{blue!5}1.4
& \cellcolor{blue!5}13.2 & \cellcolor{blue!5}3.2 \\
\midrule
\multirow{5}{*}{\makecell{Visual Pattern \\ Recognition}}
& Logic Patterns &100 & \textbf{67.9} & 3.6 & \textbf{67.9} & 3.6 & 32.1 & 3.6 & 14.3 & 7.1 & 19.1 & 3.4 & 21.4 & 5.8 \\
& Mirroring Patterns &88.9 & 40.0 & 10.0 & 45.0 & 5.0 & \textbf{50.0} & 0.0 & 25.0 & 5.0 & 0.0 & 0.0 & 20.0 & 0.0 \\
& Overlay Patterns & 100 & \textbf{50.0} & 2.9 & 44.1 & 2.9 & 20.6 & 2.9 & 32.4 & 14.7 & 13.7 & 5.6 & 25.5 & 2.8 \\
& Rotation Patterns &100 & 55.0 & 15.0 & 65.0 & 15.0 & \textbf{70.0} & 0.0 & 30.0 & 10.0 & 36.7 & 12.5 & 33.3 & 9.4 \\
\cmidrule(lr){2-15}
& \cellcolor{blue!5}\textbf{Sub-Type Overall Results} &\cellcolor{blue!5}97.8
& \cellcolor{blue!5}53.9 & \cellcolor{blue!5}1.0
& \cellcolor{blue!5}\textbf{54.9} & \cellcolor{blue!5}5.9
& \cellcolor{blue!5}37.7 & \cellcolor{blue!5}1.5
& \cellcolor{blue!5}25.5 & \cellcolor{blue!5}3.9
& \cellcolor{blue!5}17.0 & \cellcolor{blue!5}3.7
& \cellcolor{blue!5}24.8 & \cellcolor{blue!5}2.5 \\
\midrule
\multirow{1}{*}{\makecell{All}}
& \cellcolor{myyellow}\textbf{Overall Results} & \cellcolor{myyellow}94.1
& \cellcolor{myyellow}\textbf{49.7} & \cellcolor{myyellow}2.6
& \cellcolor{myyellow}34.4 & \cellcolor{myyellow}1.7
& \cellcolor{myyellow}30.2 & \cellcolor{myyellow}0.3
& \cellcolor{myyellow}19.2 & \cellcolor{myyellow}0.6
& \cellcolor{myyellow}14.2 & \cellcolor{myyellow}1.2
& \cellcolor{myyellow}16.2 & \cellcolor{myyellow}1.3 \\
\bottomrule
\end{tabular}
}
\caption{\textbf{Performance (Avg@3) of Close Source MLLMs on BabyVision.} The best results for each question type are marked in \textbf{bold}. Reported values represent the average Pass@1 accuracy across three random runs, accompanied by the standard deviation. All models are in thinking model with highest reasoning budget.}
\label{tab:main_results}
\end{table*}

\paragraph{Open-source Model Performance.}

For open-source models, the best reported system (Qwen3VL-235B-Thinking) attains 22.2\% overall, with most alternatives falling between 12\% and 19\%. Test-time thinking consistently provides measurable gains: within the Qwen3VL family, the Thinking variants outperform their Instruct counterparts (e.g., 22.2\% vs.\ 19.5\% at 235B), suggesting that explicit intermediate reasoning can partially mitigate visual uncertainty once relevant evidence is available. Nevertheless, even the largest open-source model remains far behind the top proprietary system, highlighting a persistent capability gap for the current open-source models on these fundamental visual abilities.

\begin{table*}[!t]
\renewcommand\arraystretch{1}
\setlength{\aboverulesep}{1.3pt}
\setlength{\belowrulesep}{1.3pt}
\centering
\resizebox{\linewidth}{!}{
\begin{tabular}{c | l | c c | c c | c c | c c | c c | c c}
\toprule
{\multirow{2}{*}{\textbf{Type}}} &
{\multirow{2}{*}{\textbf{Sub-Type}}} &
\multicolumn{2}{c|}{\textbf{Qwen3VL-235B-Thinking}} & \multicolumn{2}{c|}{\textbf{InternVL3.5-241B}} &
\multicolumn{2}{c|}{\textbf{GLM4.6V}} & \multicolumn{2}{c|}{\textbf{MimoVL-7B-RL}} & \multicolumn{2}{c|}{\textbf{Step3}} & \multicolumn{2}{c}{\textbf{KimiVL-A3B}} \\
\cmidrule(lr){3-4} \cmidrule(lr){5-6} \cmidrule(lr){7-8} \cmidrule(lr){9-10} \cmidrule(lr){11-12} \cmidrule(lr){13-14}
& & \textbf{Avg ($\mu$) $\uparrow$} & \textbf{Std ($\sigma$) $\downarrow$} & 
\textbf{Avg ($\mu$) $\uparrow$} & \textbf{Std ($\sigma$) $\downarrow$} & 
\textbf{Avg ($\mu$) $\uparrow$} & \textbf{Std ($\sigma$) $\downarrow$} & 
\textbf{Avg ($\mu$) $\uparrow$} & \textbf{Std ($\sigma$) $\downarrow$} & 
\textbf{Avg ($\mu$) $\uparrow$} & \textbf{Std ($\sigma$) $\downarrow$} &
\textbf{Avg ($\mu$) $\uparrow$} & \textbf{Std ($\sigma$) $\downarrow$} \\
\midrule
\multirow{9}{*}{\makecell{Fine-grained \\ Discrimination}}
& 2D Pattern Completion & 25.0 & 10.8 & 11.7 & 6.2 & \textbf{40.0} & 10.8 & 30.0 & 10.8 & 33.3 & 9.4 & 25.0 & 8.2 \\
& Count Clusters & \textbf{27.8} & 0.0 & 20.4 & 9.4 & 20.4 & 2.6 & 11.1 & 4.5 & 9.3 & 6.9 & 3.7 & 2.6 \\
& Count Same Patterns& \textbf{15.2} & 3.6 & 10.5 & 2.7 & 6.7 & 2.7 & 9.5 & 1.4 & 3.8 & 2.7 & 1.9 & 1.4 \\
& Find the Different& 16.7 & 3.0 & \textbf{20.8} & 3.0 & 14.6 & 3.0 & 2.1 & 3.0 & 0.0 & 0.0 & 6.3 & 0.0 \\
& Find the Same & \textbf{23.5} & 8.3 & 9.8 & 5.6 & 9.8 & 2.8 & 7.8 & 2.8 & 2.0 & 2.8 & 3.9 & 5.6 \\
& Reconstruction & 33.3 & 3.4 & \textbf{38.1} & 3.4 & 26.2 & 3.4 & 9.5 & 3.4 & 11.9 & 8.9 & 14.3 & 5.8 \\
& Find the Shadow & 15.9 & 5.4 & \textbf{17.4} & 3.6 & 7.3 & 2.1 & 7.3 & 2.1 & 10.1 & 2.1 & 7.3 & 5.4 \\
& Pattern and Color Completion& \textbf{38.3} & 9.4 & 31.7 & 6.2 & 25.0 & 7.1 & 35.0 & 7.1 & 21.7 & 6.2 & 13.3 & 8.5 \\
\cmidrule(lr){2-14}
 \cellcolor{white} & \cellcolor{blue!5}\textbf{Sub-Type Overall Results}& \cellcolor{blue!5}\textbf{23.3} & \cellcolor{blue!5}0.9 & \cellcolor{blue!5}18.6 & \cellcolor{blue!5}2.3 & \cellcolor{blue!5}17.4 & \cellcolor{blue!5}2.9 & \cellcolor{blue!5}14.1 & \cellcolor{blue!5}2.7 & \cellcolor{blue!5}11.3 & \cellcolor{blue!5}2.3 & \cellcolor{blue!5}8.8 & \cellcolor{blue!5}1.5 \\
\midrule

\multirow{6}{*}{\makecell{Visual Tracking}}
& Connect the Lines & 5.3 & 4.3 & 15.8 & 4.3 & 14.0 & 2.5 & 3.5 & 2.5 & 7.0 & 2.5 & \textbf{24.6} & 2.5 \\
& Lines Observation & \textbf{7.4} & 5.2 & 0.0 & 0.0 & 0.0 & 0.0 & 0.0 & 0.0 & 0.0 & 0.0 & 0.0 & 0.0 \\
& Maze & 11.7 & 6.2 & \textbf{23.3} & 2.4 & 21.7 & 2.4 & 8.3 & 2.4 & 21.7 & 2.4 & \textbf{23.3} & 4.7 \\
& Metro Map & \textbf{13.9} & 3.9 & 2.8 & 3.9 & 0.0 & 0.0 & \textbf{13.9} & 10.4 & 2.8 & 3.9 & 5.6 & 3.9 \\
& Recognize Numbers and Letters & 36.2 & 7.4 & 39.1 & 3.6 & 30.4 & 6.2 & 14.5 & 2.1 & \textbf{53.6} & 4.1 & 8.7 & 3.6 \\
\cmidrule(lr){2-14}
 \cellcolor{white} & \cellcolor{blue!5}\textbf{Sub-Type Overall Results} & \cellcolor{blue!5}16.9 & \cellcolor{blue!5}4.3 & \cellcolor{blue!5}20.5 & \cellcolor{blue!5}2.0 & \cellcolor{blue!5}16.9 & \cellcolor{blue!5}1.7 & \cellcolor{blue!5}8.8 & \cellcolor{blue!5}2.1 & \cellcolor{blue!5}\textbf{22.1} & \cellcolor{blue!5}1.5 & \cellcolor{blue!5}14.5 & \cellcolor{blue!5}2.0 \\
\midrule

\multirow{6}{*}{\makecell{Spatial Perception}} 
& 3D Cube Unfold & \textbf{19.4} & 7.9 & 8.3 & 6.8 & 8.3 & 6.8 & 8.3 & 6.8 & 11.1 & 10.4 & 2.8 & 3.9 \\
& 3D Pattern Completion & 29.6 & 6.9 & 22.2 & 4.5 & 33.3 & 7.9 & \textbf{37.0} & 2.6 & 25.9 & 9.4 & 31.5 & 14.6 \\
& 3D Views & \textbf{29.6} & 8.0 & 18.5 & 3.0 & 22.2 & 3.0 & 23.5 & 1.8 & 9.9 & 4.6 & 19.8 & 4.6 \\
& Count 3D Blocks & \textbf{10.6} & 2.1 & 6.1 & 2.1 & 6.1 & 4.3 & \textbf{10.6} & 4.3 & 3.0 & 2.1 & 6.1 & 4.3 \\
& Paper Folding & 8.3 & 6.8 & 11.1 & 3.9 & 16.7 & 0.0 & 13.9 & 3.9 & 11.1 & 7.9 & \textbf{19.4} & 3.9 \\
\cmidrule(lr){2-14}
 \cellcolor{white} & \cellcolor{blue!5}\textbf{Sub-Type Overall Results}& \cellcolor{blue!5}\textbf{20.9} & \cellcolor{blue!5}2.7 & \cellcolor{blue!5}13.9 & \cellcolor{blue!5}2.9 & \cellcolor{blue!5}18.0 & \cellcolor{blue!5}1.4 & \cellcolor{blue!5}19.8 & \cellcolor{blue!5}0.9 & \cellcolor{blue!5}11.7 & \cellcolor{blue!5}1.4 & \cellcolor{blue!5}16.5 & \cellcolor{blue!5}2.4 \\
\midrule

\multirow{5}{*}{\makecell{Visual Pattern \\ Recognition}} 
& Logic Patterns & 19.1 & 6.7 & \textbf{26.2} & 3.4 & 11.9 & 3.4 & 9.5 & 3.4 & 14.3 & 0.0 & 4.8 & 3.4 \\
& Mirroring Patterns & \textbf{26.7} & 9.4 & 23.3 & 4.7 & 23.3 & 4.7 & \textbf{26.7} & 4.7 & 6.7 & 4.7 & 6.7 & 9.4 \\
& Overlay Patterns & \textbf{33.3} & 7.3 & 25.5 & 2.8 & 13.7 & 7.3 & 21.6 & 2.8 & 31.4 & 2.8 & 21.6 & 14.0 \\
& Rotation Patterns & 40.0 & 0.0 & \textbf{43.3} & 12.5 & 33.3 & 4.7 & 26.7 & 12.5 & 16.7 & 4.7 & 16.7 & 9.4 \\
\cmidrule(lr){2-14}
\cellcolor{white} & \cellcolor{blue!5}\textbf{Sub-Type Overall Results}& \cellcolor{blue!5}\textbf{29.4} & \cellcolor{blue!5}5.8 & \cellcolor{blue!5}28.8 & \cellcolor{blue!5}1.9 & \cellcolor{blue!5}19.0 & \cellcolor{blue!5}2.5 & \cellcolor{blue!5}20.3 & \cellcolor{blue!5}3.3 & \cellcolor{blue!5}19.0 & \cellcolor{blue!5}0.9 & \cellcolor{blue!5}13.1 & \cellcolor{blue!5}4.9 \\
\midrule
\multirow{1}{*}{\makecell{All}} 
\cellcolor{white} & \cellcolor{myyellow}\textbf{Overall Results} & \cellcolor{myyellow}\textbf{22.2} & \cellcolor{myyellow}1.0 & \cellcolor{myyellow}19.2 & \cellcolor{myyellow}0.7 & \cellcolor{myyellow}17.6 & \cellcolor{myyellow}1.8 & \cellcolor{myyellow}15.1 & \cellcolor{myyellow}1.3 & \cellcolor{myyellow}14.7 & \cellcolor{myyellow}0.8 & \cellcolor{myyellow}12.4 & \cellcolor{myyellow}1.7 \\
\bottomrule
\end{tabular}
}
\caption{\textbf{Performance (Avg@3) of Open Source MLLMs on BabyVision.} The best results for each question type are marked in \textbf{bold}. Reported values represent the average Pass@1 accuracy across three random runs, accompanied by the standard deviation.}
\label{tab:main_results}
\end{table*}

\paragraph{Impact of Model Scaling.}
We further investigate how model size and reasoning strategy affect performance within the Qwen3VL family as shown in Table~\ref{tab:main_results_sizes}. The Thinking variant consistently outperforms the Instruct variant at the same scale (e.g., 22.2\% vs.\ 19.5\% at 235B), confirming that explicit intermediate reasoning provides measurable gains on visual tasks. Across model sizes, we observe that performance generally improves with scale: 235B-Thinking (22.2\%) outperforms 32B-Thinking (17.4\%), which in turn exceeds 8B-Thinking (13.1\%). However, the 4B-Thinking model (14.6\%) slightly exceeds 8B-Thinking, suggesting that scaling alone does not guarantee monotonic improvement on these fundamental perceptual tasks. These results indicate that improvements in early-vision competencies may require architectural or training innovations beyond simply increasing model parameters.

\paragraph{Comparison with Young Humans.}

As BabyVision-Mini is built for meaningful developmental comparison, its tasks are strictly vision-centric, minimizing language and prior-knowledge demands so that scores reflect visual reasoning rather than text-based inference. Its small size also makes it practical to complete within a single class period for young children.

Under this lens, the gap is striking as shown in the comparison figure (see Figure~\ref{fig:abs_fig}). Most frontier MLLMs perform well below the average 3-year-old, despite their PhD-level results on language benchmarks. Gemini3-Pro-Preview is the notable outlier—the only model consistently above the Age-3 band—yet it still lags typical 6-year-olds by about 20 points.

This highlights a core limitation: the issue is not solving “hard problems,” but struggling with pre-language visual primitives—the early perceptual and spatial abilities humans acquire before language becomes the main reasoning tool.

\paragraph{Fine-grained Analysis.}

To identify which visual primitives are most challenging for current MLLMs, we analyze performance at the subtype level across all four domains.

1. Fine-grained Discrimination. Within this category, models show highly variable performance across subtypes. Gemini3-Pro-Preview achieves relatively strong results on \emph{Reconstruction} (67.9\%) and \emph{Count Clusters} (66.7\%), but all models struggle with \emph{Find the Same} (best: 26.5\%) and \emph{Find the Different} (best: 56.3\%). These tasks require detecting subtle visual differences among highly similar patterns---a capability that demands high-fidelity perceptual encoding rather than semantic abstraction.

2. Visual Tracking. This category exposes a fundamental weakness in maintaining continuous identity across spatial trajectories. The \emph{Lines Observation} task is particularly diagnostic: Gemini3-Pro-Preview achieves 83.3\%, but most other models score near zero, indicating a complete inability to trace curves through intersections. Similarly, \emph{Connect the Lines} and \emph{Maze} tasks reveal that models frequently ``switch tracks'' at crossings---errors that are immediately obvious to human observers.

3. Spatial Perception. The \emph{Count 3D Blocks} subtype is uniformly challenging across all models, with the best accuracy at only 20.5\% (Gemini3-Pro-Preview). This task requires inferring hidden volume and maintaining a coherent 3D mental model---capabilities that cannot be verbalized without information loss. In contrast, \emph{3D Views} (66.7\% for Gemini3-Pro-Preview) proves more tractable, likely because the viewpoint transformation can be partially reasoned through geometric rules.

4. Visual Pattern Recognition. Models show comparatively stronger performance in this category, with both Gemini3-Pro-Preview and GPT-5.2 achieving 67.9\% on \emph{Logic Patterns} and Doubao-1.8 reaching 70.0\% on \emph{Rotation Patterns}. These results suggest that when patterns involve discrete, rule-based transformations (rotation, reflection), models can partially leverage their symbolic reasoning capabilities. However, \emph{Overlay Patterns} (best: 50.0\%) and \emph{Mirroring Patterns} (best: 50.0\%) remain challenging compared to the near-perfect human baselines (100\% and 88.9\% respectively), indicating that continuous spatial transformations are still problematic.

\subsection{\datasetnamegen Results}
\label{sec:gen_results}

We evaluate three frontier visual generation models on \datasetnamegen: NanoBanana-Pro, GPT-Image-1.5, and Qwen-Image-Edit. Results are presented in Table~\ref{tab:gen_results}. Note that since \datasetnamegen uses a different evaluation methodology (visual output assessment via image generation) and covers a subset of tasks adapted for generative evaluation, the results are not directly comparable to the MLLM experiments on \datasetname.

\begin{table*}[!t]
\renewcommand\arraystretch{1}
\setlength{\aboverulesep}{1.3pt}
\setlength{\belowrulesep}{1.3pt}
\centering
\resizebox{\linewidth}{!}{
\begin{tabular}{c | l | c | c c | c c | c c}
\toprule
{\multirow{2}{*}{\textbf{Type}}} &
{\multirow{2}{*}{\textbf{Sub-Type}}} &
{\multirow{2}{*}{\textbf{\#}}} &
\multicolumn{2}{c|}{\textbf{NanoBanana-Pro}} &
\multicolumn{2}{c|}{\textbf{GPT-Image-1.5}} &
\multicolumn{2}{c}{\textbf{Qwen-Image-Edit-2511}} \\
\cmidrule(lr){4-5} \cmidrule(lr){6-7} \cmidrule(lr){8-9}
& & &
\textbf{Avg ($\mu$) $\uparrow$} & \textbf{Std ($\sigma$) $\downarrow$} &
\textbf{Avg ($\mu$) $\uparrow$} & \textbf{Std ($\sigma$) $\downarrow$} &
\textbf{Avg ($\mu$) $\uparrow$} & \textbf{Std ($\sigma$) $\downarrow$} \\
\midrule
\multirow{9}{*}{\makecell{Fine-grained \\ Discrimination}}
& 2D Pattern Completion & 19 & \textbf{24.6} & 4.8 & 12.3 & 2.5 & 14.0 & 6.6 \\
& Count Clusters & 9 & \textbf{22.2} & 9.1 & 3.7 & 5.2 & 7.4 & 5.2 \\
& Count Same Patterns & 31 & \textbf{31.2} & 6.1 & 0.0 & 0.0 & 2.2 & 3.0 \\
& Find the Different & 16 & \textbf{35.4} & 2.9 & 12.5 & 8.8 & 2.1 & 2.9 \\
& Find the Same & 10 & 3.3 & 4.7 & \textbf{3.3} & 4.7 & 0.0 & 0.0 \\
& Reconstruction & 13 & \textbf{30.8} & 6.3 & 10.3 & 3.6 & 0.0 & 0.0 \\
& Find the Shadow & 14 & \textbf{9.5} & 3.4 & 7.1 & 5.8 & 2.4 & 3.4 \\
& Pattern and Color Completion & 16 & 22.9 & 2.9 & \textbf{31.3} & 0.0 & 8.3 & 5.9 \\
\cmidrule(lr){2-9}
& \cellcolor{blue!5}\textbf{Category Average} & \cellcolor{blue!5}128 & \cellcolor{blue!5}\textbf{24.5} & \cellcolor{blue!5}1.8 & \cellcolor{blue!5}9.6 & \cellcolor{blue!5}1.2 & \cellcolor{blue!5}4.7 & \cellcolor{blue!5}0.8 \\
\midrule
\multirow{5}{*}{\makecell{Visual Tracking}}
& Connect the Lines & 12 & 0.0 & 0.0 & 0.0 & 0.0 & 0.0 & 0.0 \\
& Maze & 18 & 0.0 & 0.0 & 0.0 & 0.0 & 0.0 & 0.0 \\
& Metro Map & 12 & \textbf{11.1} & 3.9 & 5.6 & 3.9 & 0.0 & 0.0 \\
& Recognize Numbers and Letters & 13 & \textbf{17.9} & 3.6 & 5.1 & 3.6 & 0.0 & 0.0 \\
\cmidrule(lr){2-9}
& \cellcolor{blue!5}\textbf{Category Average} & \cellcolor{blue!5}55 & \cellcolor{blue!5}\textbf{6.7} & \cellcolor{blue!5}2.1 & \cellcolor{blue!5}2.4 & \cellcolor{blue!5}1.0 & \cellcolor{blue!5}0.0 & \cellcolor{blue!5}0.0 \\
\midrule
\multirow{6}{*}{\makecell{Spatial Perception}}
& 3D Cube Unfold & 12 & 2.8 & 3.9 & 0.0 & 0.0 & 0.0 & 0.0 \\
& 3D Pattern Completion & 18 & 5.6 & 0.0 & \textbf{14.8} & 2.6 & 13.0 & 5.2 \\
& 3D Views & 19 & \textbf{21.1} & 8.6 & 17.5 & 2.5 & 10.5 & 4.5 \\
& Count 3D Blocks & 5 & \textbf{26.7} & 9.4 & 20.0 & 0.0 & 0.0 & 0.0 \\
& Paper Folding & 5 & \textbf{20.0} & 16.3 & 6.7 & 9.4 & 0.0 & 0.0 \\
\cmidrule(lr){2-9}
& \cellcolor{blue!5}\textbf{Category Average} & \cellcolor{blue!5}59 & \cellcolor{blue!5}\textbf{13.0} & \cellcolor{blue!5}6.0 & \cellcolor{blue!5}12.4 & \cellcolor{blue!5}2.6 & \cellcolor{blue!5}7.3 & \cellcolor{blue!5}3.5 \\
\midrule
\multirow{5}{*}{\makecell{Visual Pattern \\ Recognition}}
& Logic Patterns & 11 & \textbf{30.3} & 8.6 & 12.1 & 8.6 & 12.1 & 11.2 \\
& Mirroring Patterns & 3 & 0.0 & 0.0 & \textbf{44.4} & 15.7 & 11.1 & 15.7 \\
& Overlay Patterns & 15 & \textbf{24.4} & 16.5 & 8.9 & 2.9 & 0.0 & 0.0 \\
& Rotation Patterns & 9 & 18.5 & 5.2 & \textbf{25.9} & 5.2 & 14.8 & 10.5 \\
\cmidrule(lr){2-9}
& \cellcolor{blue!5}\textbf{Category Average} & \cellcolor{blue!5}38 & \cellcolor{blue!5}\textbf{22.8} & \cellcolor{blue!5}10.0 & \cellcolor{blue!5}16.7 & \cellcolor{blue!5}5.5 & \cellcolor{blue!5}7.9 & \cellcolor{blue!5}2.6 \\
\midrule
\multirow{1}{*}{\makecell{All}}
& \cellcolor{myyellow}\textbf{Overall} & \cellcolor{myyellow}280 & \cellcolor{myyellow}\textbf{18.3} & \cellcolor{myyellow}2.2 & \cellcolor{myyellow}9.8 & \cellcolor{myyellow}1.0 & \cellcolor{myyellow}4.8 & \cellcolor{myyellow}0.7 \\
\bottomrule
\end{tabular}
}
\caption{\textbf{Performance (\%) of Visual Generation Models on \datasetnamegen.} The best results for each subtype are marked in \textbf{bold}. \# denotes the number of questions in each subtype. Reported values represent the average accuracy across three random runs, accompanied by the standard deviation.}
\label{tab:gen_results}
\end{table*}

\paragraph{Overall Performance.}
NanoBanana-Pro achieves the highest overall accuracy at 18.3\% ($\pm$2.2), substantially outperforming GPT-Image-1.5 (9.8\% $\pm$1.0) and Qwen-Image-Edit (4.8\% $\pm$0.7). The relatively low standard deviations across runs suggest consistent behavior, though overall accuracy remains limited.

The performance distribution across categories reveals distinct patterns. NanoBanana-Pro achieves its strongest results on \emph{Fine-grained Discrimination} (24.5\%) and \emph{Visual Pattern Recognition} (22.8\%), while \emph{Spatial Perception} (13.0\%) shows moderate performance with higher variance ($\pm$6.0). In contrast, \emph{Visual Tracking} remains extremely challenging (6.7\% for the best model), with all models struggling on maze navigation and line tracing tasks. Notably, tasks requiring continuous trajectory maintenance (\emph{Maze}, \emph{Connect the Lines}) yield 0\% accuracy across all generation models, indicating that current architectures cannot maintain spatial coherence over extended generation sequences.

\paragraph{Subtype Analysis.}
Several subtypes reveal the characteristics of visual generation approaches. On \emph{Find the Different}, NanoBanana-Pro achieves 35.4\% ($\pm$2.9), and \emph{Count Same Patterns} reaches 31.2\% ($\pm$6.1). These tasks share a common characteristic: solutions are naturally expressed by drawing or marking on the image. For \emph{Spatial Perception} tasks such as \emph{Count 3D Blocks} (26.7\%) and \emph{Paper Folding} (20.0\%), NanoBanana-Pro demonstrates non-trivial performance, though with considerable variance across runs. In \emph{Visual Pattern Recognition}, the models show mixed results: \emph{Logic Patterns} (30.3\%) and \emph{Overlay Patterns} (24.4\%) yield moderate accuracy, while \emph{Mirroring Patterns} proves challenging (0\% for NanoBanana-Pro, though GPT-Image-1.5 achieves 44.4\% with high variance). These results suggest that visual generation offers a promising but currently limited pathway for visual reasoning tasks.

\section{Discussion}
\label{sec:discussion}

In this section, we discuss the implications of these findings and outline potential directions for future research.

\subsection{Why Do Frontier Models Fail on Seemingly Simple Tasks?} \label{sec: Error Analysis }

\begin{figure}[!htbp]
\vspace{-3mm}
\fontsize{8.8pt}{\baselineskip}\selectfont
\linespread{0.9}\selectfont

\begin{minipage}[t]{0.49\textwidth}
\begin{mybody}
    \centering
    \includegraphics[width=\linewidth,height=3.2cm,keepaspectratio]{images/challenge1.png}
    \begin{minipage}[t][5.5cm][t]{\linewidth}

    \textbf{Question:} Find the missing piece that fits into the empty white space within the hexagonal structure. \\
    \textbf{Answer:} \textcolor{green!60!black}{B} \quad \textbf{Model Answer:} \textcolor{red}{D} \\
    \textbf{Reasoning:} Bottom island implies split legs; check right then left flank. \\
    \textcolor{red!60!black}{Option A has a hexagon sticking out to the right... Option B also has a "shoulder"... Option C has a straight right edge... Option D has an empty right side (a narrow "neck")...} \\
    \textcolor{red!60!black}{Option C is flat on the left... Option B is compact... Option D has a left protrusion...} \\
    \textbf{Why wrong:} Over-verbalized geometry; misses the exact contour.
    \end{minipage}
\end{mybody}
\caption*{(a) Non-verbal fine-grained perception failure}
\end{minipage}
\hfill
\begin{minipage}[t]{0.49\textwidth}
\begin{mybody}
    \centering
    \includegraphics[width=\linewidth,height=3.2cm,keepaspectratio]{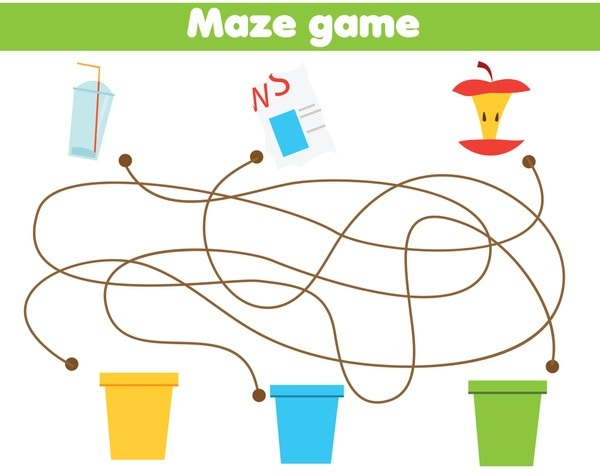}
    \begin{minipage}[t][5.5cm][t]{\linewidth}

    \textbf{Question:} Which colored trash bin will the plastic cup, old newspaper, and apple core each go into? \\
    \textbf{Answer:} \textcolor{green!60!black}{A-blue, B-yellow, C-green} \\ \textbf{Model Answer:} \textcolor{red}{A-green, B-yellow, C-blue} \\
    \textbf{Reasoning:} For plastic cup (A), start at its dot; \textcolor{red!60!black}{Follow the line to the left. It curves down and forms the outermost path on the left side of the maze.} It continues down, runs along the bottom, and goes up to green (A-green). For apple core (C), the top path goes left, down the second vertical line, and into blue (C-blue). The remaining line links the newspaper (B) to yellow. \\
    \textbf{Why wrong:} Track-switching at crossings swaps endpoints.
    \end{minipage}
\end{mybody}
\caption*{(b) Failure of manifold identity tracking}
\end{minipage}

\vspace{2mm}

\begin{minipage}[t]{0.49\textwidth}
\begin{mybody}
    \centering
    \includegraphics[width=\linewidth,height=3.2cm,keepaspectratio]{images/challenge3.png}
    \begin{minipage}[t][5.5cm][t]{\linewidth}

    \textbf{Question:} Which is the correct view when looking at the 3D structure from the direction indicated by the arrow? \\
    \textbf{Answer:} \textcolor{green!60!black}{A} \quad \textbf{Model Answer:} \textcolor{red}{C} \\
    \textbf{Reasoning:} Arrow indicates a right-side view; project the silhouette by columns. Columns 1--2 are tallest (tower behind the seat), column 3 is lower (seat), and column 4 is the step. \textcolor{red!60!black}{Its height is 2. There is nothing taller behind it. Therefore, the silhouette height is 2.} So the profile becomes 4, 4, 3, 2 and the model picks C. \\
    \textbf{Why wrong:} Verbal height logic ignores the true 3D structure and occlusion.
    \end{minipage}
\end{mybody}
\caption*{(c) Breakdown in 3D spatial imagination}
\end{minipage}
\hfill
\begin{minipage}[t]{0.49\textwidth}
\begin{mybody}
    \centering
    \includegraphics[width=\linewidth,height=3.2cm,keepaspectratio]{images/challenge4.png}
    \begin{minipage}[t][5.5cm][t]{\linewidth}

    \textbf{Question:} Among the three given options, choose the most suitable one to fill in the question mark, so that it shows a certain pattern. \\
    \textbf{Answer:} \textcolor{green!60!black}{C} \quad \textbf{Model Answer:} \textcolor{red}{B} \\
    \textbf{Reasoning:} \textcolor{red!60!black}{Each octagon is composed of 8 equal segments, with two green, two brown, and four orange segments.} The brown pair shifts two positions counterclockwise each step, so the next should place brown at the top; B matches that. Option C is rejected because brown appears at \textcolor{red!60!black}{the top left and bottom}. \\
    \textbf{Why wrong:} Focuses on color placement rather than the actual structural transform.
    \end{minipage}
\end{mybody}
\caption*{(d) Failure in abstract visual pattern induction}
\end{minipage}

\vspace{-2mm}
\caption{
Four classic vision-centric challenges for MLLMs.
All examples highlight failures caused by compressing perceptual reasoning into language.
}
\label{fig:unspeakable_challenges}
\end{figure}

Our evaluation reveals a striking inversion: state-of-the-art MLLMs that excel at PhD-level reasoning benchmarks struggle with visual tasks that young children solve effortlessly. Qualitative analysis characterizes four systematic failure modes where the visual modality gap acts as an information bottleneck. Figure~\ref{fig:unspeakable_challenges} illustrates four representative failure cases from Gemini3-Pro-Preview that motivate the analysis below.

\paragraph{Loss of Fine-Grained Detail.} A pervasive weakness across BabyVision is the degradation of fine-grained visual information, where MLLMs fail to distinguish candidates relying on sub-semantic cues like specific curvature or pixel-level boundary alignment. While humans solve these tasks via direct shape matching—a parallel geometric operation that verifies congruence without intermediate description—MLLMs rely on implicit semantic compression, attempting to discretize continuous shapes into lossy linguistic tokens. This process creates a resolution gap where fine spatial structure is flattened into semantic space, rendering micro-differences indistinguishable and forcing the model to reason through a low-fidelity proxy rather than end-to-end perceptual comparison.

\paragraph{Loss of Manifold Identity.} We observe a critical failure in topological consistency, where MLLMs struggle to maintain the identity of a continuous manifold (e.g., a winding line) as it interacts with others. Unlike humans, who utilize primitive contour integration to "lock onto" a curve and track it through occlusions, MLLMs attempt to map these 1D manifolds into discrete instruction sequences. Without a persistent, distinct representation of the specific curve, the model faces combinatorial branching at every intersection; consequently, it often "switches tracks" or hallucinates endpoints, revealing an inability to separate overlapping signals or preserve perceptual identity across extended spatial trajectories.

\paragraph{Failure of Spatial Imagination.} A third bottleneck is the inability to perform mental affine transformations—constructing and manipulating a stable 3D mental model from 2D input (e.g., rotation, projection). While humans solve these tasks via non-verbal mental simulation that retains geometric fidelity, MLLMs attempt to approximate 3D states through descriptive logic: summarizing components and inferring views via language. This fails because language acts as an insufficient coordinate system for volumetric constraints; by substituting physics-compliant rendering with probabilistic text generation, models inevitably hallucinate hidden structures or deduce impossible projections.

\paragraph{Failure of Visual Pattern Induction.} The fourth challenge is abstract rule acquisition, or the ability to induce generalized transformation rules from sparse examples. Humans readily disentangle abstract relational structures (such as rotation or nesting) from specific visual attributes, but MLLMs frequently conflate style with structure. By approaching induction through attribute enumeration rather than relational mapping, models fixate on spurious surface correlations (e.g., specific colors) instead of the underlying transformation logic, failing to treat objects as variables in a logical operation and missing the compositional generalization required for these tasks.

\paragraph{The Verbalization Bottleneck}

A unifying theme across all four failure modes is what we term the \emph{verbalization bottleneck}. Current MLLMs process visual inputs by translating them into language representations before reasoning. While this approach leverages the powerful reasoning capabilities of large language models, it introduces a fundamental limitation: visual information that cannot be faithfully expressed in language is lost.

Consider the distinction between \emph{describable} and \emph{undescribable} visual properties. Semantic content (``a red car on a road'') translates well to language, but geometric relationships (the precise curvature of a boundary, the exact position of an intersection point) resist verbalization. BabyVision specifically targets these undescribable properties, which explains why models that excel at language-heavy benchmarks fail here.

This analysis suggests that progress on early-vision tasks may require architectural innovations that preserve visual information throughout the reasoning process, rather than compressing it into a linguistic bottleneck.

\subsection{Insight from Training: Can RLVR Help Visual Reasoning?}
\label{subsec:training}

We have identified large performance gaps in BabyVision, not only among human and frontier models, but also across closed- and open-source models. We are further interested in: how can we develop stronger visual reasoning skills and achieve better scores on BabyVision with open models?

As Reinforcement Learning with Verifiable Rewards (RLVR) has recently demonstrated strong gains in language-reasoning performance for LLMs, we conduct an init study to investigate whether RLVR can similarly improve visual abilities measured by BabyVision. We use Qwen3-VL-8B-Thinking as the base model and apply RLVR fine-tuning. For data collection, we adopt a BabyVision-style pipeline but draw from larger image sources and remove duplicates, yielding 1,400 training examples. The collected data covers all four major BabyVision task families, yet its difficulty distribution is not completely aligned with BabyVision: the model achieves 34.2\% initial accuracy on the RLVR training set, but only 13.1\% on BabyVision, when evaluated with the same base model.

\begin{figure}[t]
\centering
\includegraphics[width=1.0\textwidth]{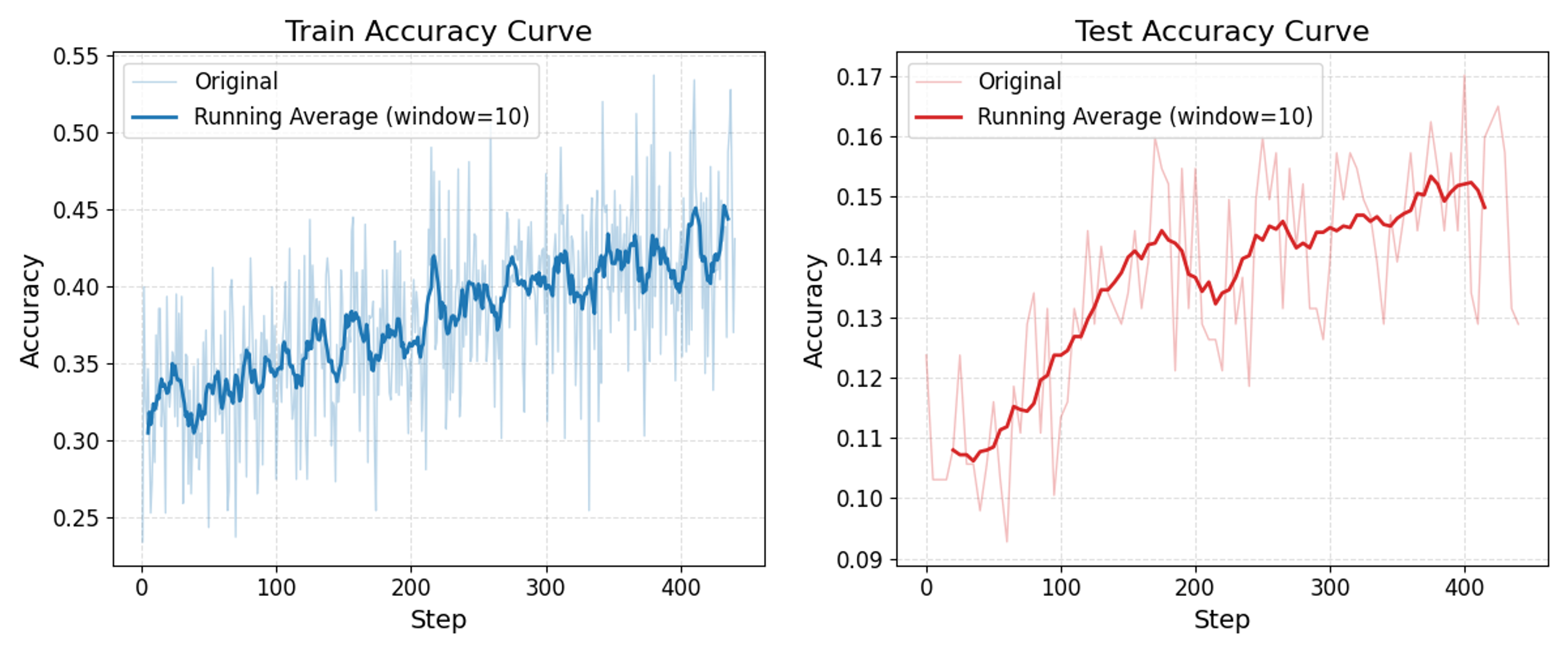}
\caption{GRPO training dynamics for Qwen3-VL-8B-Thinking. Both training accuracy and held-out test accuracy steadily improve during training.}
\label{fig:rl_curve}
\end{figure}

\paragraph{Training Details.} We train Qwen3-VL-8B-Thinking using GRPO on 8 H800 GPUs for approximately 3 days over 18 epochs. To allow for sufficient exploration, we configure the rollout n to 10, the max response length to 16384, and set the clip range (clip higher) to 0.28. Both the global batch size and rollout batch size are set to 64 within the EasyR1~\citep{zheng2025easyr1,sheng2024hybridflow} framework. We employ an LLM judge to determine the consistency between the answer extracted from \texttt{\textbackslash boxed\{\}} label and the ground truth, using it as the reward signal to guide the model. We observe that RLVR is effective on the collected training dataset: both training accuracy and held-out test accuracy consistently improve over the course of training as shown in Figure~\ref{fig:rl_curve}.

\begin{figure}[t]
\centering
\includegraphics[width=0.7\textwidth]{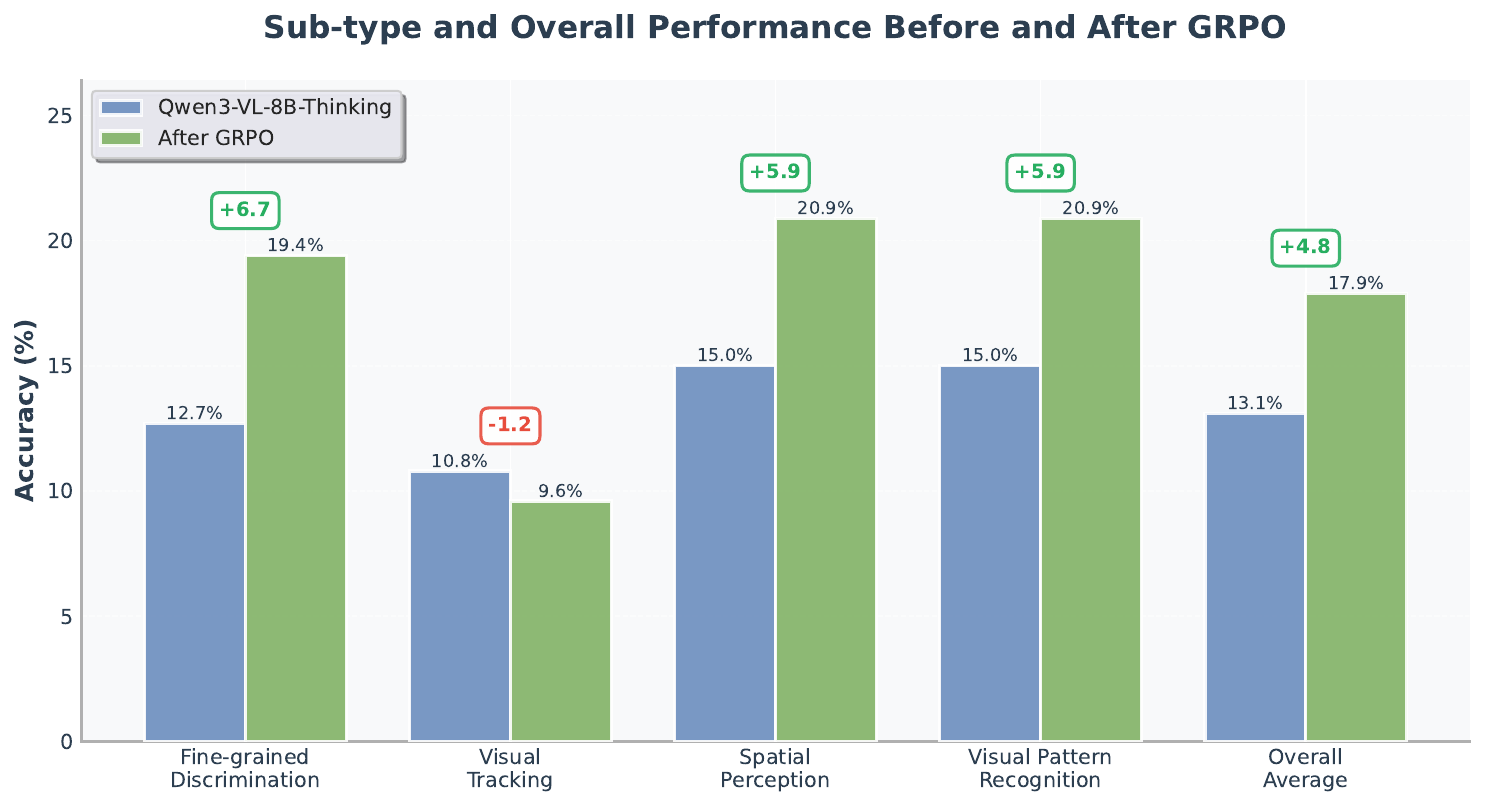}
\caption{BabyVision accuracy before vs.\ after RLVR fine-tuning. RLVR yields a +4.8 overall accuracy gain and improves most subtypes.}
\label{fig:rl_bar}
\end{figure}

\paragraph{Results and Analysis.} The BabyVision performance of Qwen3-VL-8B-Thinking before and after RL fine-tuning is reported in Figure~\ref{fig:rl_bar}. The model achieves a +4.8-point overall accuracy improvement after RLVR training. We also observe consistent gains across most task subtypes, with the sole exception of visual tracking, for which RL fine-tuning yields little to even negative improvement. We think that this is because visual tracking is the least amenable to verbalization; since RLVR primarily enhances performance by encouraging longer and more structured “thinking-token” reasoning, it provides less benefit on tasks that depend on continuous perceptual tracking rather than language-mediated reasoning. Detailed performance comparison is listed in Table~\ref{tab:grpo_subtype_final} in the Appendix.






\begin{figure}[!t]
\centering
\includegraphics[width=0.85\textwidth]{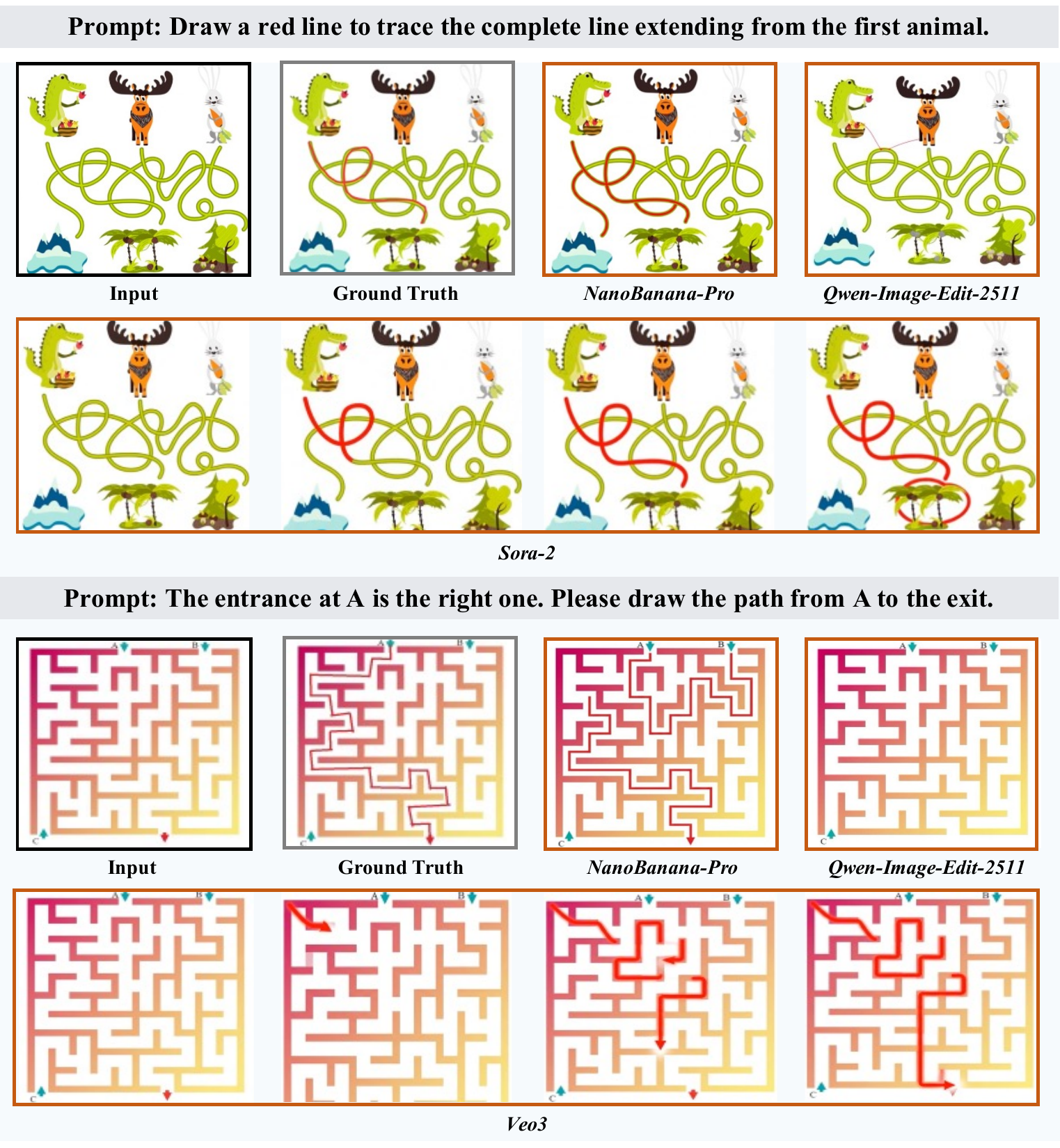}
\caption{Representative examples of visual reasoning results from different image/video generation models evaluated on \datasetnamegen.}
\label{fig:vis_externalization}
\end{figure}

\subsection{Beyond Language: Visual Externalization}

A key observation motivating \datasetnamegen is that many BabyVision tasks are most naturally solved by \emph{visual externalization}---drawing, marking, or tracing on the image. When humans solve maze or line-tracking problems, they often point or draw trajectories rather than verbalizing directions. This suggests an alternative paradigm: instead of reasoning in language space and outputting a verbal answer, models could reason in visual space and output a visual solution.

Our experiments with generation models (image and video) on \datasetnamegen reveal promising signals. As shown in Figure~\ref{fig:vis_externalization}, we present examples of visual reasoning results from generation models\footnote{More output images and videos are available at \url{https://unipat.ai/blog/BabyVision}}. Models like Sora-2 and NanoBanana-Pro exhibit human-like visual thinking processes, explicitly drawing trajectories along paths. However, their solutions still contain noticeable errors, indicating that visual generation alone is insufficient---the generation process must be guided by robust visual understanding.

This points to a compelling research direction: \emph{visual generation models as multimodal reasoners}. If models can learn to manipulate images directly---tracing paths, completing patterns, transforming shapes---they may bypass the verbalization bottleneck entirely. Recent unified multimodal models that natively integrate visual understanding and generation, such as Bagel~\citep{bagel2025}, offer a new path forward. Unlike traditional MLLMs that compress visual inputs through a language bottleneck, these natively multimodal architectures maintain visual representations throughout the reasoning process, enabling them to explicitly think in visual space---sketching intermediate steps, highlighting regions, or drawing solution trajectories. Similarly, frontier video generation models like Sora 2~\citep{sora2} and Veo 3~\citep{veo3} demonstrate emergent capabilities in modeling physical dynamics and spatial relationships, allowing the generative process itself to serve as a form of visual reasoning~\citep{MMGR}.

\section{Conclusion}

We introduce BabyVision, a benchmark for evaluating early-vision abilities that humans develop before language acquisition. Our evaluation of state-of-the-art MLLMs reveals a striking gap: even the best model (Gemini3-Pro-Preview, 49.7\%) falls far short of human performance (94.1\%), with consistent deficits across fine-grained discrimination, visual tracking, spatial perception, and pattern recognition. Error analysis identifies four failure modes---loss of fine-grained detail, loss of manifold identity, failure of spatial imagination, and confusion of appearance with structure---all stemming from a verbalization bottleneck that discards undescribable visual information.

These findings suggest that strong performance on language-heavy benchmarks does not imply robust visual foundations. Progress will likely require architectural innovations that preserve visual fidelity throughout reasoning, rather than compressing perception into language. We also introduce \datasetnamegen, a generative extension enabling evaluation through visual outputs, and show preliminary evidence that generation models can exhibit human-like visual thinking. BabyVision provides a diagnostic tool for measuring progress toward multimodal systems with genuinely grounded visual intelligence.

\section{Acknowledgements}
We would like to thank Xiaotao Gu (Zhipu AI), Junyang Lin (Alibaba Group), Shuai Bai (Alibaba Group), and Shuhuai Ren (Xiaomi MiMo) for their valuable discussions and insightful feedback throughout this project.

\clearpage
\appendix
\newpage
\appendix
\onecolumn

\section{Evaluation Details for \datasetname}
\label{appx:eval}

We employ Qwen3-Max as the judge model to evaluate semantic equivalence between model outputs and ground-truth answers. The judge receives the following prompt:
\begin{tcolorbox}[colback=gray!5!white, colframe=gray!75!black, title=EVALUATION PROMPT FOR \datasetname]
\small\ttfamily
You are a careful and strict evaluator. You will be given:

1. QUESTION

2. GROUND TRUTH ANSWER (correct answer)

3. MODEL OUTPUT (answer from another model)

YOUR GOAL: Determine if the Model Output \textbf{accurately matches} the Ground Truth Answer in meaning.

MATCHING MEANS: the facts, entities, and key details are equivalent, even if phrasing differs.

NOT MATCHING MEANS: the Model Output is wrong, incomplete, contains extra incorrect facts, or changes the meaning.

PROCESS (INTERNAL REASONING):
\begin{enumerate}[nosep]
\item Read and understand the Question, Ground Truth Answer, and Model Output.
\item Ignore small wording differences, formatting, or synonyms.
\item If all factual content matches, conclude True. Otherwise, conclude False.
\end{enumerate}

IMPORTANT: Think through your decision step-by-step \textbf{internally} before responding. In your final output, return \textbf{only} True or False, with no extra text or explanation.

OUTPUT FORMAT: True or False

INPUT:\\
Question: \{Question\}\\
Ground Truth Answer: \{Groundtruth\}\\
Model Output: \{Model Output\}
\end{tcolorbox}

\section{Evaluation Details for \datasetnamegen}
\label{appx:eval-gen}

\subsection{Inference Protocol}

For \datasetnamegen, we instruct visual generation models to annotate the original image with their solution. This ensures that the original visual context is preserved while allowing models to demonstrate their reasoning through visual marks. We use the following prompt template:

\begin{tcolorbox}[colback=blue!3!white, colframe=blue!60!black, title=\textbf{Generation Prompt Template}]
\small\ttfamily
\textbf{CRITICAL INSTRUCTION:} You are a visual annotation assistant. Your task is to add ONLY the requested annotations (circles, lines, arrows, text labels) to mark the answer on the image.

\textbf{IMPORTANT RULES:}
\begin{enumerate}[nosep]
\item DO NOT modify, redraw, or alter ANY part of the original image content
\item DO NOT change colors, shapes, positions, or any visual elements of the original image
\item ONLY add overlay annotations (circles, lines, arrows, text) on TOP of the original image
\item The original image must remain 100\% intact and unchanged
\item Use bright, visible colors (red, green, blue) for your annotations so they stand out
\item Keep annotations minimal and precise - only mark what is asked
\end{enumerate}

\textbf{YOUR TASK:} \{Question\}

\textbf{REMINDER:} Keep the original image EXACTLY as it is. ONLY add annotation marks (circles, lines, arrows, or text) to indicate your answer. Do not redraw or modify any part of the original image.
\end{tcolorbox}

\subsection{Automatic Evaluation Protocol}

We employ Gemini-3-Flash as the judge model to evaluate whether generated images correctly solve the visual reasoning task. The evaluation is conducted by comparing three images: (1) the original input image, (2) the ground-truth solution image annotated by human experts, and (3) the model-generated image. The judge determines whether the generated solution matches the ground truth.

We design type-specific evaluation criteria for each subtype to ensure accurate assessment. The general prompt structure is:

\begin{tcolorbox}[colback=green!3!white, colframe=green!60!black, title=\textbf{Automatic Evaluation Prompt}]
\small\ttfamily
You are evaluating an AI-generated image for a visual reasoning task.

\textbf{TASK TYPE:} \{task\_type\}\\
\textbf{SUBTYPE:} \{subtype\}\\
\textbf{GENERATION INSTRUCTION:} "\{generation\_prompt\}"

\textbf{You are provided with THREE images:}
\begin{itemize}[nosep]
\item Image 1 (Input): The original question/puzzle image
\item Image 2 (Ground Truth): The CORRECT answer showing what the result SHOULD look like
\item Image 3 (Generated): The AI-generated result to be evaluated
\end{itemize}

Compare Image 3 (Generated) with Image 2 (Ground Truth) to determine if they show the SAME answer.

\textbf{[Type-Specific Criteria Inserted Here]}

\textbf{DECISION RULES:}
\begin{itemize}[nosep]
\item TRUE: Generated image shows the EXACT SAME answer as Ground Truth
\item FALSE: Generated shows a DIFFERENT answer, NO answer, or UNCLEAR answer
\end{itemize}

\textbf{IMPORTANT:}
\begin{itemize}[nosep]
\item Focus ONLY on whether the ANSWER matches, ignore style differences
\item A marking on a DIFFERENT element/option = FALSE
\item A path taking a DIFFERENT route = FALSE
\item A DIFFERENT number or character = FALSE
\item Missing required answer = FALSE
\end{itemize}

\textbf{Respond with ONLY one word:} "True" or "False"
\end{tcolorbox}

\subsection{Type-Specific Evaluation Criteria}

We customize evaluation criteria for each task subtype. The complete criteria are provided below:

\subsubsection{Fine-grained Discrimination}

\begin{tcolorbox}[colback=orange!3!white, colframe=orange!60!black, title=\textbf{Find the Different}]
\small\ttfamily
\textbf{TASK:} Find the unique/different element among many similar elements.\\
\textbf{CRITERIA:} Is the circle on the SAME grid position as ground truth?\\
- Circle on DIFFERENT element = FALSE\\
- No circle visible = FALSE
\end{tcolorbox}

\begin{tcolorbox}[colback=orange!3!white, colframe=orange!60!black, title=\textbf{Find the Same}]
\small\ttfamily
\textbf{TASK:} Find identical elements or matching figures.\\
\textbf{CRITERIA:} Are ALL marked elements the same as in ground truth?\\
- Circles on DIFFERENT elements = FALSE\\
- Missing circles = FALSE
\end{tcolorbox}

\begin{tcolorbox}[colback=orange!3!white, colframe=orange!60!black, title=\textbf{Find the Shadow}]
\small\ttfamily
\textbf{TASK:} Find the shadow/silhouette that matches the colored figure.\\
\textbf{CRITERIA:} Is the SAME option circled?
\end{tcolorbox}

\begin{tcolorbox}[colback=orange!3!white, colframe=orange!60!black, title=\textbf{Reconstruction / 2D Pattern Completion / Pattern and Color Completion}]
\small\ttfamily
\textbf{TASK:} Find/select the correct option.\\
\textbf{CRITERIA:} Is the SAME option circled/selected?
\end{tcolorbox}

\begin{tcolorbox}[colback=orange!3!white, colframe=orange!60!black, title=\textbf{Count Same Patterns / Count Clusters}]
\small\ttfamily
\textbf{TASK:} Count patterns or fill in numbers.\\
\textbf{CRITERIA:} Do the markings/numbers match ground truth exactly?
\end{tcolorbox}

\subsubsection{Visual Tracking}

\begin{tcolorbox}[colback=purple!3!white, colframe=purple!60!black, title=\textbf{Maze}]
\small\ttfamily
\textbf{TASK:} Draw a path through the maze.\\
\textbf{CRITERIA:} Does the path follow the EXACT SAME route as ground truth?\\
- Different route = FALSE\\
- No visible path = FALSE
\end{tcolorbox}

\begin{tcolorbox}[colback=purple!3!white, colframe=purple!60!black, title=\textbf{Connect the Lines}]
\small\ttfamily
\textbf{TASK:} Trace a line following the continuous path.\\
\textbf{CRITERIA:} Does the traced line follow the SAME path as ground truth?
\end{tcolorbox}

\begin{tcolorbox}[colback=purple!3!white, colframe=purple!60!black, title=\textbf{Metro Map}]
\small\ttfamily
\textbf{TASK:} Draw the shortest path between metro stations.\\
\textbf{CRITERIA:} Does the path follow the EXACT SAME route as ground truth?
\end{tcolorbox}

\begin{tcolorbox}[colback=purple!3!white, colframe=purple!60!black, title=\textbf{Recognize Numbers and Letters}]
\small\ttfamily
\textbf{TASK:} Fill in letters/numbers in blanks.\\
\textbf{CRITERIA:} Are the EXACT SAME characters filled in each blank?
\end{tcolorbox}

\subsubsection{Spatial Perception}

\begin{tcolorbox}[colback=teal!3!white, colframe=teal!60!black, title=\textbf{3D Views / 3D Cube Unfold / Paper Folding / 3D Pattern Completion}]
\small\ttfamily
\textbf{TASK:} Select the correct option for spatial reasoning.\\
\textbf{CRITERIA:} Is the SAME option circled?
\end{tcolorbox}

\begin{tcolorbox}[colback=teal!3!white, colframe=teal!60!black, title=\textbf{Count 3D Blocks}]
\small\ttfamily
\textbf{TASK:} Count cubes in a 3D structure.\\
\textbf{CRITERIA:} Is the EXACT SAME number written?
\end{tcolorbox}

\subsubsection{Visual Pattern Recognition}

\begin{tcolorbox}[colback=red!3!white, colframe=red!60!black, title=\textbf{Logic Patterns / Mirroring Patterns / Overlay Patterns / Rotation Patterns}]
\small\ttfamily
\textbf{TASK:} Identify pattern and select correct option.\\
\textbf{CRITERIA:} Is the SAME option circled?
\end{tcolorbox}

\subsection{\datasetnamegen Statistics}
\label{appx:gen-stats}

Table~\ref{tab:gen_statistics} presents the detailed distribution of questions across categories and subtypes in \datasetnamegen.

\begin{table}[h]
\centering
\renewcommand\arraystretch{1.1}
\begin{tabular}{l l r}
\toprule
\textbf{Category} & \textbf{Subtype} & \textbf{\#} \\
\midrule
\multirow{8}{*}{Fine-grained Discrimination}
& 2D Pattern Completion & 19 \\
& Count Clusters & 9 \\
& Count Same Patterns & 31 \\
& Find the Different & 16 \\
& Find the Same & 10 \\
& Reconstruction & 13 \\
& Find the Shadow & 14 \\
& Pattern and Color Completion & 16 \\
\cmidrule{2-3}
& \textit{Subtotal} & \textit{128} \\
\midrule
\multirow{4}{*}{Visual Tracking}
& Connect the Lines & 12 \\
& Maze & 18 \\
& Metro Map & 12 \\
& Recognize Numbers and Letters & 13 \\
\cmidrule{2-3}
& \textit{Subtotal} & \textit{55} \\
\midrule
\multirow{5}{*}{Spatial Perception}
& 3D Cube Unfold & 12 \\
& 3D Pattern Completion & 18 \\
& 3D Views & 19 \\
& Count 3D Blocks & 5 \\
& Paper Folding & 5 \\
\cmidrule{2-3}
& \textit{Subtotal} & \textit{59} \\
\midrule
\multirow{4}{*}{Visual Pattern Recognition}
& Logic Patterns & 11 \\
& Mirroring Patterns & 3 \\
& Overlay Patterns & 15 \\
& Rotation Patterns & 9 \\
\cmidrule{2-3}
& \textit{Subtotal} & \textit{38} \\
\midrule
\multicolumn{2}{l}{\textbf{Total}} & \textbf{280} \\
\bottomrule
\end{tabular}
\caption{Distribution of questions across categories and subtypes in \datasetnamegen. One subtype from \datasetname is excluded as it does not naturally admit a generative solution format.}
\label{tab:gen_statistics}
\end{table}

\subsection{Human Evaluation and Validation}

To validate the reliability of our automatic evaluation pipeline, we conducted comprehensive human evaluation on all 280 NanoBanana-Pro outputs. PhD-level annotators independently judged each generated image against the ground truth.

\paragraph{Overall Agreement.} The automatic scorer achieves \textbf{96.1\%} agreement with human judgments (269/280 samples). The confusion matrix is shown in Table~\ref{tab:gen_confusion}.

\begin{table}[h]
\centering
\renewcommand\arraystretch{1.1}
\begin{tabular}{l|cc|c}
\toprule
& \textbf{Human: Correct} & \textbf{Human: Incorrect} & \textbf{Total} \\
\midrule
\textbf{Auto: Correct} & \cellcolor{green!15}67 & \cellcolor{red!15}3 & 70 \\
\textbf{Auto: Incorrect} & \cellcolor{red!15}8 & \cellcolor{green!15}202 & 210 \\
\midrule
\textbf{Total} & 75 & 205 & 280 \\
\bottomrule
\end{tabular}
\caption{Confusion matrix for automatic vs. human evaluation on NanoBanana-Pro outputs. Green cells indicate agreement; red cells indicate disagreement.}
\label{tab:gen_confusion}
\end{table}

\paragraph{Evaluation Metrics.} The key metrics are summarized below:
\begin{itemize}[nosep]
\item \textbf{Human Evaluation Accuracy:} 75/280 (26.8\%)
\item \textbf{Auto Evaluation Accuracy:} 70/280 (25.0\%)
\item \textbf{Precision} (auto correct when human correct): 0.957
\item \textbf{Recall} (finds human correct cases): 0.893
\item \textbf{F1 Score:} 0.924
\end{itemize}

\section{Additional Experimental Results}

\begin{table*}[h]
\renewcommand\arraystretch{1}
\setlength{\aboverulesep}{1.3pt}
\setlength{\belowrulesep}{1.3pt}
\centering
\resizebox{\linewidth}{!}{
\begin{tabular}{c | l | c c | c c | c c | c c | c c}
\toprule
{\multirow{2}{*}{\textbf{Type}}} &
{\multirow{2}{*}{\textbf{Sub-Type}}} &
\multicolumn{2}{c|}{\textbf{Qwen3VL-235B-Thinking}} & \multicolumn{2}{c|}{\textbf{Qwen3VL-235B-Instruct}} &
\multicolumn{2}{c|}{\textbf{Qwen3VL-32B-Thinking}} & \multicolumn{2}{c|}{\textbf{Qwen3VL-8B-Thinking}} & \multicolumn{2}{c}{\textbf{Qwen3VL-4B-Thinking}} \\
\cmidrule(lr){3-4} \cmidrule(lr){5-6} \cmidrule(lr){7-8} \cmidrule(lr){9-10} \cmidrule(lr){11-12}
& & \textbf{Avg ($\mu$) $\uparrow$} & \textbf{Std ($\sigma$) $\downarrow$} & 
\textbf{Avg ($\mu$) $\uparrow$} & \textbf{Std ($\sigma$) $\downarrow$} & 
\textbf{Avg ($\mu$) $\uparrow$} & \textbf{Std ($\sigma$) $\downarrow$} & 
\textbf{Avg ($\mu$) $\uparrow$} & \textbf{Std ($\sigma$) $\downarrow$} & 
\textbf{Avg ($\mu$) $\uparrow$} & \textbf{Std ($\sigma$) $\downarrow$} \\
\midrule
\multirow{9}{*}{\makecell{Fine-grained \\ Discrimination}}
& 2D Pattern Completion        & 25.0 & 10.8 & 31.7 & 10.3 & \textbf{38.3} & 4.7 & 23.3 & 8.5 & 20.0 & 8.2 \\
& Count Clusters              & 27.8 & 0.0  & 16.7 & 4.5  & \textbf{35.2} & 5.2 & 20.4 & 2.6  & 16.7 & 4.5 \\
& Count Same Patterns         & \textbf{15.2} & 3.6  & 11.4 & 0.0  & 11.4 & 2.3 & 2.9  & 2.3  & 3.8  & 1.4 \\
& Find the Different          & \textbf{16.7} & 3.0  & 4.2  & 3.0  & 6.3  & 5.1 & 0.0  & 0.0  & 0.0  & 0.0 \\
& Find the Same               & \textbf{23.5} & 8.3  & 11.8 & 4.8  & 7.8  & 2.8 & 3.9  & 2.8  & 11.8 & 4.8 \\
& Reconstruction    & \textbf{33.3} & 3.4  & 21.4 & 5.8  & 23.8 & 6.7 & 19.1 & 3.4  & 19.1 & 3.4 \\
& Find the Shadow             & 15.9 & 5.4  & \textbf{24.6} & 2.1  & \textbf{24.6} & 5.4 & 11.6 & 2.1  & 8.7  & 6.2 \\
& Pattern and Color Completion& \textbf{38.3} & 9.4  & 10.0 & 4.1  & 25.0 & 7.1 & 26.7 & 4.7  & 23.3 & 6.2 \\
\cmidrule(lr){2-12}
 \cellcolor{white} & \cellcolor{blue!5}\textbf{Sub-Type Overall Results}& \cellcolor{blue!5}\textbf{23.3} & \cellcolor{blue!5}0.9 & \cellcolor{blue!5}16.4 & \cellcolor{blue!5}1.3 & \cellcolor{blue!5}21.1 & \cellcolor{blue!5}1.8 & \cellcolor{blue!5}12.7 & \cellcolor{blue!5}0.6 & \cellcolor{blue!5}12.1 & \cellcolor{blue!5}2.3 \\
\midrule

\multirow{6}{*}{\makecell{Visual Tracking}}
& Connect the Lines            & 5.3  & 4.3 & 10.5 & 7.4 & 10.5 & 4.3 & \textbf{12.3}  & 2.5 & 10.5 & 0.0 \\
& Lines Observation            & \textbf{7.4}  & 5.2 & 0.0  & 0.0 & 0.0  & 0.0 & 0.0  & 0.0 & 0.0  & 0.0 \\
& Maze                         & 11.7 & 6.2 & 11.7 & 4.7 & 5.0  & 4.1 & 20.0  & 0.0 & \textbf{28.3} & 6.2 \\
& Metro Map                    & 13.9 & 3.9 & 13.9 & 3.9 & \textbf{19.4} & 3.9 & 5.6  & 3.9 & 5.6  & 7.9 \\
& Recognize Numbers and Letters& \textbf{36.2} & 7.4 & 24.6 & 2.1 & 17.4 & 3.6 & 8.7 & 0.0 & 17.4 & 3.6 \\
\cmidrule(lr){2-12}
 \cellcolor{white} & \cellcolor{blue!5}\textbf{Sub-Type Overall Results} & \cellcolor{blue!5}\textbf{16.9} & \cellcolor{blue!5}4.3 & \cellcolor{blue!5}15.3 & \cellcolor{blue!5}0.6 & \cellcolor{blue!5}8.8 & \cellcolor{blue!5}2.5 & \cellcolor{blue!5}10.8 & \cellcolor{blue!5}1.0 & \cellcolor{blue!5}14.9 & \cellcolor{blue!5}1.1 \\
\midrule

\multirow{6}{*}{\makecell{Spatial Perception}} 
& 3D Cube Unfold              & \textbf{19.4} & 7.9 & 13.9 & 3.9 & 16.7 & 6.8 & 5.6  & 3.9 & 8.3  & 6.8 \\
& 3D Pattern Completion       & 29.6 & 6.9 & \textbf{31.5} & 5.2 & 24.1 & 5.2 & 22.2 & 4.5 & 25.9 & 9.4 \\
& 3D Views                    & \textbf{29.6} & 8.0 & \textbf{29.6} & 3.0 & 11.1 & 8.0 & 21.0 & 8.7 & \textbf{29.6} & 3.0 \\
& Count 3D Blocks             & 10.6 & 2.1 & \textbf{21.2} & 9.3 & 7.6  & 2.1 & 7.6 & 5.7 & 12.1 & 4.3 \\
& Paper Folding               & 8.3  & 6.8 & 11.1 & 10.4& 5.6  & 3.9 & \textbf{13.9} & 7.9 & 8.3  & 6.8 \\
\cmidrule(lr){2-12}
 \cellcolor{white} & \cellcolor{blue!5}\textbf{Sub-Type Overall Results}& \cellcolor{blue!5}20.9 & \cellcolor{blue!5}2.7 & \cellcolor{blue!5}\textbf{23.4} & \cellcolor{blue!5}2.7 & \cellcolor{blue!5}12.8 & \cellcolor{blue!5}1.9 & \cellcolor{blue!5}15.0 & \cellcolor{blue!5}3.2 & \cellcolor{blue!5}19.1 & \cellcolor{blue!5}1.4 \\
\midrule

\multirow{5}{*}{\makecell{Visual Pattern \\ Recognition}} 
& Logic Patterns              & 19.1 & 6.7 & \textbf{21.4} & 5.8 & \textbf{21.4} & 5.8 & 11.9  & 3.4 & 14.3 & 5.8 \\
& Mirroring Patterns          & 26.7 & 9.4 & 20.0 & 0.0 & \textbf{30.0} & 8.2 & 6.7  & 4.7 & 10.0 & 8.2 \\
& Overlay Patterns            & \textbf{33.3} & 7.3 & \textbf{33.3} & 2.8 & 19.6 & 5.6 & 7.8 & 5.6 & 11.8 & 0.0 \\
& Rotation Patterns           & 40.0 & 0.0 & 43.3 & 12.5& \textbf{46.7} & 9.4 & 40.0 & 14.1& 23.3 & 12.5 \\
\cmidrule(lr){2-12}
\cellcolor{white} & \cellcolor{blue!5}\textbf{Sub-Type Overall Results}& \cellcolor{blue!5}\textbf{29.4} & \cellcolor{blue!5}5.8 & \cellcolor{blue!5}\textbf{29.4} & \cellcolor{blue!5}4.2 & \cellcolor{blue!5}27.5 & \cellcolor{blue!5}3.2 & \cellcolor{blue!5}15.0 & \cellcolor{blue!5}1.9 & \cellcolor{blue!5}14.4 & \cellcolor{blue!5}2.5 \\
\midrule
\multirow{1}{*}{\makecell{All}} 
\cellcolor{white} & \cellcolor{myyellow}\textbf{Overall Results} & \cellcolor{myyellow}\textbf{22.2} & \cellcolor{myyellow}1.0 & \cellcolor{myyellow}19.5 & \cellcolor{myyellow}1.4 & \cellcolor{myyellow}17.4 & \cellcolor{myyellow}2.0 & \cellcolor{myyellow}13.1 & \cellcolor{myyellow}1.1 & \cellcolor{myyellow}14.6 & \cellcolor{myyellow}1.2 \\
\bottomrule
\end{tabular}
}
\caption{\textbf{Performance (Avg@3) of Qwen3VL Instruct/Thinking on BabyVision.} The best results for each question type are marked in \textbf{bold}. Reported values represent the average Pass@1 accuracy across three random runs, accompanied by the standard deviation.}
\label{tab:main_results_sizes}
\end{table*}

\begin{table*}[t!]
\footnotesize
\centering
\setlength{\tabcolsep}{7pt}
\renewcommand{\arraystretch}{1.12}
\resizebox{0.9\linewidth}{!}{
\begin{tabular}{
l !{\vrule width 0.8pt}
l !{\vrule width 0.8pt}
c !{\vrule width 0.8pt}
c !{\vrule width 0.8pt}
c
}
\toprule
\textbf{Category} &
\textbf{Subtype} &
\textbf{Qwen3-VL-8B-Thinking (\%)} &
\textbf{After GRPO (\%)} &
\textbf{$\Delta$ (\%)} \\
\midrule

\multirow{9}{*}{\makecell[l]{\textbf{Fine-grained}\\\textbf{Discrimination}}} &
2D Pattern Completion & 23.3 & \textbf{45.0} & \deltaCell{+21.7} \\
& Count Clusters & \textbf{20.4} & \textbf{20.4} & \deltaCell{+0.0} \\
& Count Same Patterns & 2.9 & \textbf{5.7} & \deltaCell{+2.8} \\
& Find the different & 0.0 & \textbf{2.1} & \deltaCell{+2.1} \\
& Find the same & 3.9 & \textbf{15.7} & \deltaCell{+11.8} \\
& Reconstruction & 19.1 & \textbf{23.8} & \deltaCell{+4.7} \\
& Find the shadow & 11.6 & \textbf{17.4} & \deltaCell{+5.8} \\
& Pattern and Color Completion & 26.7 & \textbf{33.3} & \deltaCell{+6.6} \\
& \cellcolor{rowgray}{\textbf{Sub-Type Overall Results}}
  & \cellcolor{rowgray}{12.7}
  & \cellcolor{rowgray}{\textbf{19.4}}
  & \cellcolor{rowgray}{\deltaCell{+6.7}} \\
\midrule

\multirow{6}{*}{\makecell[l]{\textbf{Visual}\\\textbf{Tracking}}} &
Connect the lines & \textbf{12.3} & 3.5 & \deltaCell{-8.8} \\
& Lines Observation & \textbf{0.0} & \textbf{0.0} & \deltaCell{+0.0} \\
& Maze & \textbf{20.0} & 15.0 & \deltaCell{-5.0} \\
& Metro map & 5.6 & \textbf{13.9} & \deltaCell{+8.3} \\
& Recognize numbers and letters & 8.7 & \textbf{11.6} & \deltaCell{+2.9} \\
& \cellcolor{rowgray}{\textbf{Sub-Type Overall Results}}
  & \cellcolor{rowgray}{\textbf{10.8}}
  & \cellcolor{rowgray}{9.6}
  & \cellcolor{rowgray}{\deltaCell{-1.2}} \\
\midrule

\multirow{6}{*}{\makecell[l]{\textbf{Spatial}\\\textbf{Perception}}} &
3D Cube Unfold & 5.6 & \textbf{16.7} & \deltaCell{+11.1} \\
& 3D Pattern Completion & 22.2 & \textbf{37.0} & \deltaCell{+14.8} \\
& 3D Views & 21.0 & \textbf{25.9} & \deltaCell{+4.9} \\
& Count 3D blocks & 7.6 & \textbf{9.1} & \deltaCell{+1.5} \\
& Paper Folding & \textbf{13.9} & 11.1 & \deltaCell{-2.8} \\
& \cellcolor{rowgray}{\textbf{Sub-Type Overall Results}}
  & \cellcolor{rowgray}{15.0}
  & \cellcolor{rowgray}{\textbf{20.9}}
  & \cellcolor{rowgray}{\deltaCell{+5.9}} \\
\midrule

\multirow{5}{*}{\makecell[l]{\textbf{Visual Pattern}\\\textbf{Recognition}}} &
Logic Patterns & 11.9 & \textbf{26.2} & \deltaCell{+14.3} \\
& Mirroring Patterns & 6.7 & \textbf{20.0} & \deltaCell{+13.3} \\
& Overlay Patterns & 7.8 & \textbf{13.7} & \deltaCell{+5.9} \\
& Rotation Patterns & \textbf{40.0} & 26.7 & \deltaCell{-13.3} \\
& \cellcolor{rowgray}{\textbf{Sub-Type Overall Results}}
  & \cellcolor{rowgray}{15.0}
  & \cellcolor{rowgray}{\textbf{20.9}}
  & \cellcolor{rowgray}{\deltaCell{+5.9}} \\
\midrule

\multicolumn{2}{l !{\vrule width 0.8pt}}{\textbf{Overall Average Accuracy}} &
13.1 & \textbf{17.9} & \deltaCell{+4.8} \\
\bottomrule
\end{tabular}}

\caption{GRPO results by subtype (in \%). Vertical rules separate category, subtype, and metric groups. For each row, the larger value between the original and GRPO model is bolded.}
\label{tab:grpo_subtype_final}
\vspace{-3mm}
\end{table*}

\clearpage
\bibliography{example_paper}
\bibliographystyle{colm2024_conference}

\end{document}